%% file: ms.tex
\documentclass[final]{cvpr}

\usepackage{times}
\usepackage{epsfig}
\usepackage{graphicx}
\usepackage{amsmath}
\usepackage{amssymb}
\usepackage{booktabs}       %

\usepackage[capbesideposition=outside,capbesidesep=quad]{floatrow}		%
\newfloatcommand{capbtabbox}{table}[][\FBwidth]
\usepackage{multirow}
\usepackage{arydshln}
\usepackage{comment}
\usepackage{enumitem}
\usepackage{mathtools}
\usepackage{stfloats}		%
\usepackage[usenames, dvipsnames]{xcolor}
\definecolor{blueTeaser}{RGB}{91, 155, 213}
\definecolor{greenTeaser}{RGB}{112, 173, 71}

\definecolor{bluePipeline}{RGB}{68, 114, 196}
\definecolor{greenPipeline}{RGB}{112, 173, 71}
\usepackage[pagebackref=true,breaklinks=true,colorlinks,bookmarks=false,hypertexnames=false]{hyperref}
\usepackage[numbers,sort,compress]{natbib} %

\newcommand{\boldparagraph}[1]{\vspace{0.05em}\noindent{\bf #1} }

\newcommand{\urlNewWindow}[1]{\href[pdfnewwindow=true]{#1}{\nolinkurl{#1}}}

\DeclarePairedDelimiter\floor{\lfloor}{\rfloor}

\def\cvprPaperID{4353} %
\def\confYear{CVPR 2021}

\begin{document}

\title{DeepSurfels: Learning Online Appearance Fusion}

\author{
    Marko Mihajlovic$^1$
	\qquad
	Silvan Weder$^1$
	\qquad
	Marc Pollefeys$^{1,2}$
	\qquad
	Martin R.~Oswald$^1$\\
	{\normalsize 
	$^1$Department of Computer Science, ETH Zurich \qquad
	$^2$Microsoft Mixed Reality and AI Zurich Lab}\\[4pt]
    {\small \href[pdfnewwindow=true]{https://onlinereconstruction.github.io/DeepSurfels}{\nolinkurl{OnlineReconstruction.github.io/DeepSurfels}}}
}

\maketitle
\thispagestyle{empty}

\input{tex/0_abstract}
\input{tex/1_introduction}

\input{tex/2_related_work}

\input{tex/3_deep_surfels}

\input{tex/4_appearance_fusion}

\input{tex/5_experiments}

\input{tex/6_conclusion}

\noindent
\begin{minipage}{\columnwidth}
	\vspace{8pt}
	\footnotesize
	\noindent
	\textbf{Acknowledgments.}~
	\input{tex/8_acknowledgements}
\end{minipage}

{\small
\bibliographystyle{ieee_fullname}
\bibliography{egbib}
}
\input{supplementary_material/main}

\end{document}

%% file: tex/0_abstract.tex
\begin{abstract}
We present DeepSurfels, a novel hybrid scene representation for geometry and appearance information.
DeepSurfels combines explicit and neural building blocks to jointly encode geometry and appearance information. In contrast to established representations, DeepSurfels better represents high-frequency textures, is well-suited for online updates of appearance information, and can be easily combined with machine learning methods. 
We further present an end-to-end trainable online appearance fusion pipeline that fuses information from RGB images into the proposed scene representation and is trained using self-supervision imposed by the reprojection error with respect to the input images. 
Our method compares favorably to classical texture mapping approaches as well as recent learning-based techniques. 
Moreover, we demonstrate lower runtime, improved generalization capabilities, and better scalability to larger scenes compared to existing methods.

\end{abstract}

%% file: tex/1_introduction.tex
\begin{figure*}[t]
    \centering
    \includegraphics[width=\textwidth]{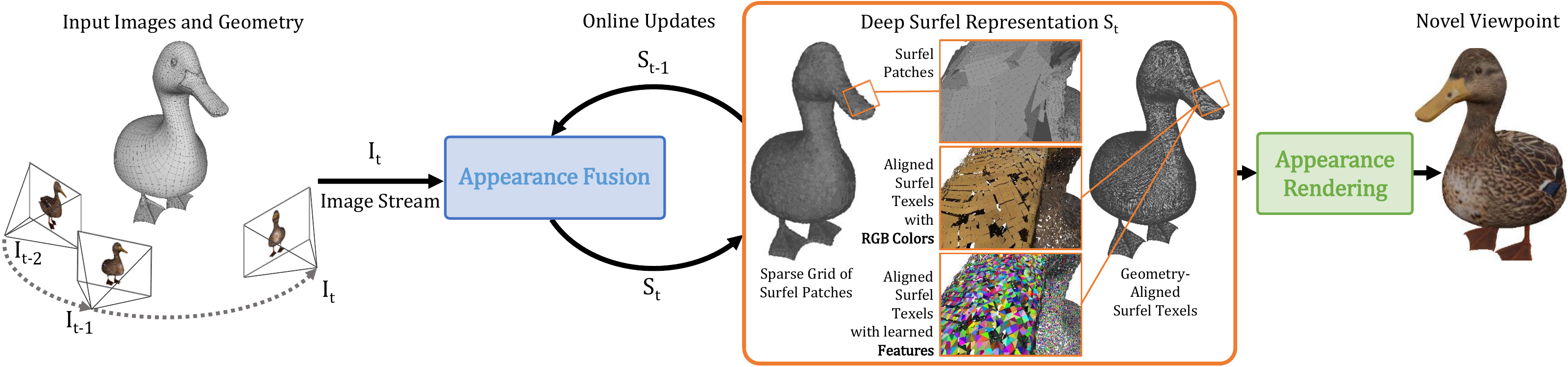}
    \vspace{-19pt}
    \caption{\textbf{Overview of our online appearance fusion pipeline and the DeepSurfel scene representation.}
    The \textbf{\textcolor{blueTeaser}{Appearance Fusion}} network efficiently aggregates appearance information from a stream of camera views into the proposed DeepSurfel representation $S_{t-1}$ that maintains high-frequency geometric and appearance information.
    DeepSurfels is a sparse grid of 2D patches that consist of surface-aligned texels, which encode appearance information either as RGB color values or learned feature vectors.
    The proposed \textbf{\textcolor{greenTeaser}{Appearance Rendering}} network interprets aggregated and interpolated geometric and appearance information stored in DeepSurfels for rendering novel viewpoints.
    In this example we used DeepSurfels with a sparse $64^3$ patch grid with $8 \times 8$ resolution surfel patches.}
    \label{fig:pipeline}
\end{figure*}

\section{Introduction}
Realistic 3D model reconstruction from images and depth sensors has been a central and long-studied problem in computer vision.
Appearance mapping is often treated as a separate post-processing step that follows 3D surface reconstruction and is usually approached using batch-based optimization methods~\cite{debevec1996modeling, Eisemann-et-al-CGF-2008, Waechter-et-al-ECCV-14, Fu-et-al-CVPR-2018} that are unsuitable for many applications that do not have access to the entire dataset at processing time, for instance, robot navigation~\cite{breitenmoser2012surface, garrido2013application, bircher2015structural}, augmented reality~\cite{newcombe2011dtam, schops2017real}, and virtual reality~\cite{Lombard-et-al-TOG-2018, lombardi2019neural,Chu-et-al-ECCV-2020} applications, Simultaneous Localization and Mapping (SLAM) systems~\cite{whelan2015real}, online scene perception methods~\cite{hane20173d, schops2017large}, and many others.

Common online fusion methods like KinectFusion~\cite{KinectFusion} are well suited for online geometry fusion and can efficiently handle noise and topological changes.
However, due to their high memory requirements at high voxel resolutions, they have strong limitations when it comes to encoding high-frequency appearance details on the surface.
On the other hand, meshes with high-resolution texture maps~\cite{Waechter-et-al-ECCV-14, Fu-et-al-CVPR-2018, Eisemann-et-al-CGF-2008} are well-suited for encoding high-frequency appearance information in an efficient manner, but they have difficulties in handling topology changes in an online reconstruction setting.
Moreover, recent learning-based approaches~\cite{Sitzmann-et-al-CVPR-2019, Sitzmann-et-al-NIPS-2019, mildenhall2020nerf, Oechsle-et-al-ICCV-2019} have achieved high-quality results by learning geometry and texture mapping directly from RGB images.
However, they are not well suited for local online updates, do not scale to large-scale scenes, and easily overfit to the training data.

In this paper, we approach the problem of online appearance reconstruction from RGB-D images by combining the advantages of
\textbf{1)} implicit grids, which easily handle topological changes and where low resolution is often sufficient to encode the scene topology, \textbf{2)} scalable high-frequency appearance along the surface via texture maps or learned feature maps, and
\textbf{3)} a learned scene representation to build a framework for learning-based appearance fusion that allows for online processing and scalability to large scenes.
To this end, we propose a novel scene representation \textbf{DeepSurfels} and an efficient learning-based \textbf{online appearance fusion pipeline} which is illustrated in \figurename~\ref{fig:pipeline}.

Our DeepSurfels representation is a hybrid between an implicit surface that encodes the topology and low-frequency geometric details and a surfel representation that encodes high-frequency geometry and appearance information in form of surface-aligned patches.
These patches are arranged in a sparse grid and consist of surface-aligned texels that encode appearance information \textit{either} in the classical form of RGB color values \textit{or}, as proposed, via learned feature vectors.
The sparse grid allows for efficient volumetric rendering %
and enables explicit scene updates that are crucial for online fusion, while the 2D patches enable quadratic memory storage complexity like meshes or sparse grid structures.
Depending on the DeepSurfel parameters it can approximate between simple colored voxels (high grid resolution, $1\times 1$ patches) and textured meshes with high texture atlas resolutions (lower grid resolution, higher patch resolution).
Our online appearance fusion pipeline iteratively fuses RGB-D frames into estimated DeepSurfels geometry and is optimized by using a differentiable renderer for self-supervision and the reprojection error as training signal.
In this way, the pipeline does not require any ground-truth texture maps and the training procedure allows for efficient transfer to new sensors and scenes without the need for acquiring costly ground-truth data.

While we eventually target full online reconstruction of both geometry and texture from monocular video, we only focus on online appearance estimation in this paper. %
Even in a setting with known geometry, our online approach has scalability advantages: We can fuse arbitrary numbers of input frames and the grid-aligned surfels have performance advantages during feature aggregation across local neighbors and for controlling the sampling density.
Our grid-aligned surfel patches can also be seen as a spatial alignment of per-voxel sub-features being anchored along the surface.
In contrast to works that only save a single feature vector per voxel, \eg DeepVoxels~\cite{Sitzmann-et-al-CVPR-2019}, we can directly relate sub-features with particular image pixels via projective mapping and as such simplify the network learning task and improve output accuracy.
As opposed to many novel view synthesis works~\cite{Sitzmann-et-al-NIPS-2019,Sitzmann-et-al-CVPR-2019,mildenhall2020nerf}, we do not overfit onto a single scene, but train a network that generalizes over multiple scenes without re-training.
While those methods iterate many times over each input image in a slow optimization process, our method processes every image only once with a single network forward pass and is thus much faster.
From the application point of view, our approach is thus closer to classical texture mapping methods like \cite{debevec1996modeling,Eisemann-et-al-CGF-2008,Fu-et-al-CVPR-2018,Waechter-et-al-ECCV-14}.

We compare our novel scene representation and appearance fusion pipeline to existing methods on single and multi-object datasets and show that our scene representation better captures high-frequency textures. %
Moreover, our method generalizes well and compares favorably even to existing texture optimization methods that jointly optimize all images together.
This is a crucial step towards a fully end-to-end appearance fusion method that can be deployed to real-world applications.
In sum, our key \textbf{contributions} are:
\begin{itemize}[itemsep=-3pt,topsep=1pt,leftmargin=*]
    \item \textbf{DeepSurfels.} A novel scalable and memory-efficient 3D scene representation closing the gap between traditional interpretable and modern learned representations.
    \item \textbf{Online Appearance Fusion Pipeline.} An end-to-end differentiable and efficient online appearance fusion pipeline compatible with classical and learned texture mapping.
    The method yields competitive texturing results without heavy optimization as every input frame is processed only \textit{once} with a single network forward pass.
    \item \textbf{Generalized Novel View Synthesis.} Contrary to other learning-based methods \cite{Sitzmann-et-al-CVPR-2019,Sitzmann-et-al-NIPS-2019,mildenhall2020nerf} that overfit onto a single scene, our method generalizes to new scenes without retraining.
\end{itemize}

%% file: tex/2_related_work.tex
\section{Related Work}

Our method relates to, and builds upon previous work on scene representations and appearance estimation which are reviewed in the following subsections.
\subsection{Scene Representations}
Scene representations can be broadly divided into explicit geometric and learned representations. 

\boldparagraph{Explicit Geometric Representations.} The major advantage of explicit geometric representations is their direct interpretability. 
\textit{Point clouds}~\cite{achlioptas2017learning, fan2017point, Park-et-al-CVPR-2019} are a lightweight and flexible 3D representation being the raw output of many 3D scanners, RGB-D cameras, and LiDARs.
However, they are less suitable for the extraction of watertight surfaces due to lacking topology and connectivity information. 
This also impedes realistic rendering with detailed textures and complex lighting.
\textit{Mesh} representations~\cite{kanazawa2018learning, wang2018pixel2mesh, liu2019soft, Hanocka2020p2m} scale well and texture mapping is convenient.
However, topological changes are difficult to handle in an online process.
\textit{Voxel-grids}~\cite{brock2016generative, gadelha20173d, liao2018deep, rezende2016unsupervised, stutz2018learning, wu2016learning} -- as a natural extension of pixels to 3D space -- easily handle topological changes but are difficult to use with textures and complex light models. Another problem arises from the cubic memory complexity of the dense representation, which makes it expensive to capture precise shape details of complex objects. 
\textit{Surface elements (Surfels)}~\cite{pfister2000surfels, Schoeps-et-al-TPAMI-2020, whelan2015elasticfusion, wang2019real} are non-connected point primitives that reduce geometry to the essentials needed for rendering, thus being more memory efficient than meshes while still providing good rendering properties. 
Our \textit{DeepSurfel} representation provides several advantages over existing explicit representations:
\textbf{1)} it maintains better connectivity information than point clouds and surfels, \textbf{2)} scales better than voxel grids while still compatible with octree~\cite{zeng2012memory, zeng2013octree, steinbrucker2013large} and voxel hashing~\cite{niessner2013real, kahler2015very} approaches that improve memory efficiency, \textbf{3)} provides better rendering quality than point clouds or voxel grids, \textbf{4)} enables fast rendering, and \textbf{5)} allows local updates. 

\boldparagraph{Learned Shape Representations.} 
Recent learned implicit representations have achieved impressive results in modeling geometry~\cite{Park-et-al-CVPR-2019,Mescheder-et-al-CVPR-2019,Chen-et-al-CVPR-2019, LEAP:CVPR:21}.
They learn a neural network to predict the signed distance or occupancy for a given query point.
These approaches struggle to scale to larger scenes and to capture high-frequency details as they tend to learn low-frequency functions which often results in over-smoothed geometry~\cite{rahaman2018spectral}.
This problem has been approach by the idea of using local features \cite{Chiyu-et-al-CVPR-2020, peng2020convolutional, Xu-et-al-NIPS-2019, chabra2020deep} or directly regressing local geometries \cite{weder2020routedfusion, weder2020neuralfusion} for improved scalability and representation power of implicit representations.
Our proposed DeepSurfels follows this direction using only local learned features for scalability. 
Unlike most learning-based methods, we do not encode the scene by network optimization. 
Instead, we train a network to estimate latent features following a data-driven approach, which better generalizes and facilitates online updates.

\subsection{Appearance Estimation}

\boldparagraph{Classical Texture Mapping.} The classical way of coloring a surface from a set of input images with known camera pose is to un-project the image information onto the surface and perform a selection or blending operation to fuse the color information~\cite{debevec1996modeling, wood2000surface, Allene-et-al-ICPR-2008}.
Due to errors in the camera alignment or in the surface geometry, blurry textures or patch seams affect results and additional texture alignment procedures have been proposed~\cite{Lensch-et-al-GM-2001, Bernardini-et-al-TVCG-2001, Theobalt-et-al-TVCG-2007, Eisemann-et-al-CGF-2008, Gal-et-al-CGF-2010, Waechter-et-al-ECCV-14, Lempitsky-Ivanov-CVPR-2007, Takai-et-al-3DPVT-2010, Fu-et-al-CVPR-2020} to tackle these problems.
Better texture mapping results have been achieved with an optical flow-like correction in texture space~\cite{Eisemann-et-al-CGF-2008, Waechter-et-al-ECCV-14, Fu-et-al-CVPR-2018}, patch-based optimization~\cite{Bi-et-al-TOG-2017}, or via 2D perspective warp techniques~\cite{lee2020texturefusion}.
With significantly more computation effort, it is also possible to better leverage the redundancy of multiple surface observations from different views and to compute super-resolved texture maps via energy minimization~\cite{Goldluecke-et-al-IJCV-2014, Tsiminaki-et-al-CVPR-2014, Fu-et-al-CVPR-2018,Tsiminaki-et-al-BMVC-2019} or with deep learning techniques~\cite{Li-et-al-CVPR-2019, Richard-et-al-3DV2019}.
All previously mentioned methods share the strategy of aggregating appearance information in patches or texture atlases with corresponding coordinates onto a mesh-based surface, while other works use voxel grids~\cite{KinectFusion, Zollhofer-et-al-TOG-2015, Maier-et-al-ICCV-2017, kutulakos1999theory, seitz1999voxelcoloring, szeliski1998stereo}, or mesh colors~\cite{Yuksel-et-al-TOG-2010, Armando-et-al-3DV-2019}. 
An overview of texture mapping methods with different representations is given in~\cite{Tarini-et-al-SIGGRAPH-Course-2017, yuksel2019rethinking}.

\boldparagraph{Learned Appearance Representations.} Recent learned appearance representations have achieved state-of-the-art results and outperformed most classical texturing methods. 
They encode visual information into learned features and store them in voxel grids \cite{flynn2019deepview, mildenhall2019local, penner2017soft, srinivasan2019pushing, lombardi2019neural, rematas2020neural}, point clouds \cite{aliev2019neural}, or meshes \cite{thies2019neural,Riegler2020FVS,zhang2020neural} which are rendered using neural networks. 
\cite{Oechsle-et-al-ICCV-2019, oechsle2020learning} use a neural network conditioned on geometry to generate a learned texture representation. 
\cite{DVR} combines geometry and appearance to generate a joint implicit representations.
Worrall~\etal~\cite{Worrall-et-al-ICCV-2017} learn a disentangled representation %
to interpret and manipulate learned feature-based scene representations.
SRNs~\cite{Sitzmann-et-al-NIPS-2019} encode the scene into a neural network and render novel views using a neural ray-tracer.
NeRF~\cite{mildenhall2020nerf} inputs the viewing direction together with point coordinates which allows to also model illumination and complex non-Lambertian surfaces. 
Other works~\cite{saito2019pifu, saito2020pifuhd} take advantage of local features for higher representation power, while~\cite{huang2020advtex, thies2020imageguided} uses appropriate loss terms to correct for geometric misalignment.
Recent trends and applications of neural renderers are summarized in~\cite{tewari2020state}.
However, these methods are currently limited to fixed-size scenes, do not scale well to larger real-world scenes, or are unsuitable for online processing of appearance information.
The global volumetric appearance reconstruction approach~\cite{Bi-et-al-ECCV-2020} additionally separates albedo, roughness, and lighting. %
Liu~\etal~\cite{Liu-et-al-TVCG-2019} present a learned approach for shape and texture reconstruction that linearly fuses shape and color information in a voxel grid as in~\cite{curless1996volumetric} and post-process the grid with a multi-resolution neural network. 
However, pure post-processing methods may not be able to revert errors of an incorrect earlier linear fusion.

\boldparagraph{Global vs Local Appearance Representations.}
Existing learned scene representations can be separated into global and local approaches. 
\cite{Oechsle-et-al-ICCV-2019, mildenhall2020nerf} are global approaches encoding the scene into a single feature vector or the weights of a neural network, while \cite{Sitzmann-et-al-CVPR-2019} can be considered a local approach that uses a dense grid of feature vectors. 
For better scalability and higher representation power, we follow the local direction. 
Further, the local storage of appearance keeps the updates of the encoded information local. 
Moreover, it allows to exploit geometric relations better constraining the learning problem for improved generalization.

\begin{figure*}[t]
    \centering
    \includegraphics[width=\textwidth]{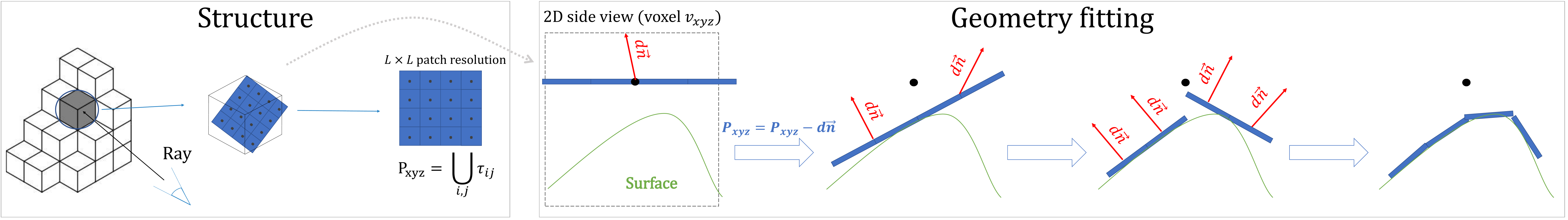}
    \vspace{-16pt}
    \caption{\textbf{DeepSurfel surface fitting.} In a recursive fitting procedure, we align the texels of each patch with the underlying SDF surface by shifting its location and adjusting its normal.
    $d$ denotes queried signed distance, $\Vec{n}$ denotes SDF gradient $\nabla \text{SDF}$ in $x, y, z$ directions. 
    }
    \label{fig:surfel_alignment}
\end{figure*}

\boldparagraph{Online Appearance Aggregation.}
Most texture mapping methods process all input images in a batch-based way after the geometry estimation step and are implemented as a separate post-processing step, whereas only a minority addresses the problem of online appearance reconstruction.
A popular work is KinectFusion~\cite{KinectFusion} and related works~\cite{Zollhofer-et-al-TOG-2015, Maier-et-al-ICCV-2017, lee2020texturefusion}, which estimate surface and appearance information from a stream of RGB-D images.
Other works fuse both geometry and appearance information directly into an oriented surfel cloud \cite{Schoeps-et-al-TPAMI-2020, whelan2015elasticfusion, wang2019real}.
The major drawback of these methods is limited capacity to store high-frequency appearance along the surface and low-quality renderings.
Therefore, we propose an efficient online appearance estimation pipeline mitigating these limitations. %

%% file: tex/3_deep_surfels.tex
\section{DeepSurfels 3D Scene Representation}
We propose \textit{DeepSurfels} as a powerful, scalable, and easy-to-use alternative to mitigate previously mentioned problems of many scene representations.

\boldparagraph{Data Structure.} DeepSurfels is a set of patches with $L \times L$ texels that can either store color information or learned feature vectors.
The elementary building block is an oriented texel $\tau \in \mathbb{R}^c$ that is associated with its weight parameter $\omega$ and is stored on the objects' surface, where $c$ denotes the number of feature channels.
This number can be chosen arbitrarily for learned appearance fusion as suited for the problem setting, while we set $c=3$ for deterministic RGB texturing.
The texels $\tau$ are arranged in an $L \times L$ resolution patch $P_{xyz}:~\{i,j \rightarrow \tau_{ij};\ i, j \in [1, L] \}$ that is located in a sparse patch grid $\mathcal{P} = \{P_{xyz}\}_{x \leq X, y \leq Y, z \leq Z}$, where $X, Y, Z$ represent DeepSurfels' grid resolution. 
Although the spatial patch size can be chosen arbitrarily, we empirically observed that texturing works best when the patch size is equal to the grid cell size such that there is no overlap between neighboring patches. 
For efficiency reasons, it is sufficient to store patches only for grid cells that intersect the objects' surface.
However, it is also possible to allocate more layers around the iso-surface to account for noisy geometry as it is common for geometric fusion approaches~\cite{KinectFusion}.

\boldparagraph{Surface Fitting.} We propose a recursive algorithm to align each texel $\tau_{ij}$ of the patch with the implicit surface of the geometry.
We compute the patch position and orientation from a signed distance function (SDF) representing the Euclidean distance to the closest surface. %

Initially, every patch $P_{xyz}$ in the grid $\mathcal{P}$ is positioned at the center of its grid cell. 
Then, the patches are shifted to the closest surface by using the pre-computed SDF, oriented according to the SDF gradient $\nabla \text{SDF}$ in all $x, y, z$ directions, and rotated to maximize the surface coverage.
These patches are subdivided into $\kappa^2$ non-overlapping patches of $\frac{L}{\kappa} \times \frac{L}{\kappa}$ resolution, where $\kappa \geq 2$ is the smallest integer to non-trivially divide $L$.
Each sub-patch is aligned again using the SDF field, where we trilinearly interpolate the SDF value at non-integer grid positions.
This patch subdivision and alignment is repeated recursively until the resolution reaches $1 \times 1$ when patches represent texels that lie on the isosurface. This process is visually illustrated in \figurename~\ref{fig:surfel_alignment}.

%% file: tex/4_appearance_fusion.tex
\section{Online Appearance Fusion Pipeline} \label{sec:fusion}

\begin{figure*}
    \centering
    \includegraphics[width=\textwidth]{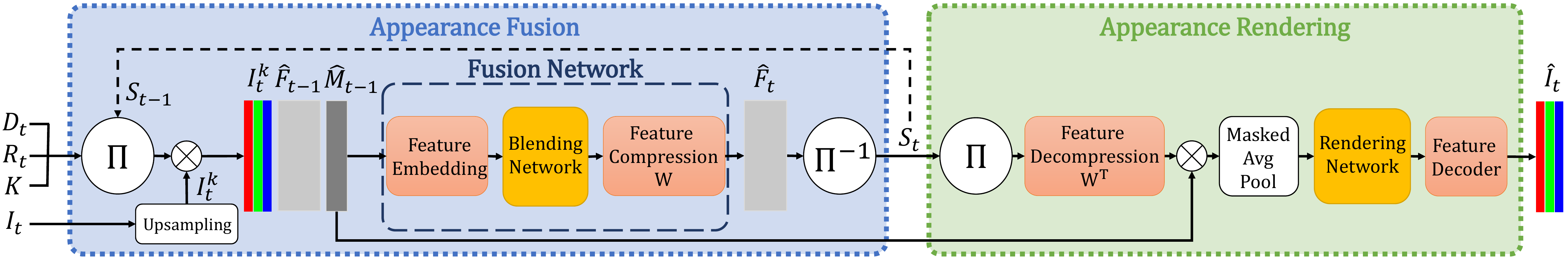}
    \caption{\textbf{Overview of our learned appearance fusion pipeline.} 
    The pipeline consists of an \textbf{\textcolor{bluePipeline}{Appearance Fusion}} module that integrates a new RGB measurement $I_t$ into DeepSurfels $S_{t-1}$ and a differentiable \textbf{\textcolor{greenPipeline}{Appearance Rendering}} module that interprets and renders the content of representation for a given viewpoint.
    White blocks denote differentiable deterministic operations, rectangular blocks denote data, rounded rectangular blocks are trainable modules, and $\otimes$ is a feature stacking operation.}
    \vspace{-2pt}
    \label{fig:full_pipeline}
\end{figure*}
We also propose a pipeline for learning appearance fusion (depicted in \figurename~\ref{fig:full_pipeline}) that incrementally fuses RGB measurements into DeepSurfels at every time step $t$ and yields DeepSurfel state $S_t$. 
The input to our pipeline are intrinsic $K_t$ and extrinsic $R_t$ camera parameters, an RGB image $I_t \in \mathbb{R}^{H \times W \times C}$, and corresponding depth map $D_t \in \mathbb{R}^{H \times W}$, where $H$, $W$ and $C$ denote image height, width, and the number of channels respectively. 
The pipeline consists of four main components detailed in the following.

\boldparagraph{Differentiable Projection $\Pi$.} The projection module renders a super-resolved feature map $\hat{F}_{t-1} \!\!\in\!\! \mathbb{R}^{k H \times k W \times c}$, where $k$ is an upsampling factor inspired by \cite{Cook-et-al-SIGGRAPH-1984} to ensure dense coverage of the geometry.
There are three steps to render this feature map from already stored scene content.

\underline{First}, each pixel in the incoming frame $D_t$ is subdivided into $k^2$ distinct sub-pixels $p_{ij}^t$ ($i \in [1, k H]$, $j \in [1, k W]$), thus forming an upsampled image grid. %

\underline{Second}, by leveraging camera and depth information, the center of the sub-pixel $p_{ij}^t$ is un-projected into the scene. 
From the un-projected scene point, the closest DeepSurfel texel and all texels within the surrounding $l_\infty$ ball are selected. 
The size of this ball is chosen proportional to the size of the un-projected sub-pixel in the world space.

\underline{Third}, an efficient uniform average of the selected texels determines the value of the feature entry $\hat{f}_{ij}^{t-1}\!\in\! \hat{F}_{t-1}$ (\ref{eq:local_feature}):
\begin{equation} \label{eq:local_feature}
    \hat{f}_{ij}^{t-1} \!=\! \frac{1}{|T_{ij}^t|}\sum_{\tau \in T_{ij}^t}{\tau}\,,
\end{equation}
where $T_{ij}^t$ is the set of selected texels. 
This algorithm is simple, leverages the grid representation for fast rendering, and can flexibly render further optionally stored features or a surface normal map $\hat{N}_{t-1}$ that we jointly denote as meta features $\hat{M}_{t-1}$.
Note that all operations are differentiable and the selection can be implemented as a differentiable multiplication by an indicator function.

\boldparagraph{Fusion Network.}
The input image $I_t$ is deterministically upsampled $I_t^k \in \mathbb{R}^{k H \times k W \times C}$ by factor $k$ (nearest-neighbor interpolation) and stacked $\otimes$ with the super-resolved features $\hat{F}_{t-1} \otimes \hat{M}_{t-1} \otimes I_t^k$. 
This stacked representation is embedded into a higher-dimensional feature space by a trainable linear transformation (\textit{Feature Embedding} module \figurename~\ref{fig:full_pipeline}).
Then, the embeddings are refined by \textit{Blending Network} that consists of five convolutional layers ($3 \times 3$ kernel size) interleaved with dropout and leaky ReLU activations. 
This network, based on a small receptive field, produces refined features aware of neighboring information that alleviates the problem of discretization artifacts, which can occur for low DeepSurfels resolutions. 
Lastly, these features are compressed by \textit{Feature Compression} $W$ layer to a lower dimensional feature space that is defined by DeepSurfels' number of channels. 
The final output is an updated feature map $\hat{F}_t$ that blends old information from $\hat{F}_{t-1}$ with the new appearance information from $I_t$.

\boldparagraph{Inverse Projection $\Pi^{-1}$.}
While the fusion module and the explicit geometry representation preserve spatial coherence, this module is responsible for integrating the new appearance information in a temporally coherent way. 
Without temporal coherence, a new observation could overwrite old states minimizing the reprojection error for the current frame while erasing valuable prior information. 
The inverse projection module $\Pi^{-1}$ integrates the updated feature map $\hat{F}_t$ into the representation $S_{t-1}$ to produce the new state $S_t$. 
For efficiency reasons, only texel values $\forall \tau_{t-1} \in \bigcup_{i,j}^{kH,kW} T^{t-1}_{i,j}$ and their weights $\omega_{t-1}$ that were intersected by at least one of the sub-pixels are updated using the following moving average scheme:
\begin{align} \label{eq:update}
    \tau_t &= \frac{1}{\omega_{t-1}\!+\!1} \Bigg(\!
        \tau_{t-1} \omega_{t-1} + \frac{\sum_{i,j}^{kH, kW} \hat{f}^{t}_{ij} \; \mathbb{I}_{ \tau_{t-1} \in T^{t-1}_{ij}}}
                     {\sum_{i,j}^{kH, kW} \mathbb{I}_{ \tau_{t-1} \in T^{t-1}_{ij}}} \!\Bigg), \nonumber \\
    \omega_{t} &= \omega_{t-1} \!+\! 1,
\end{align}
where $\mathbb{I}_E$ is an indicator function being one, if $E$ is true, and zero otherwise.
The texel weights are initialized to $\omega_{0}=0$.

The new state $S_t$ is optimally computed in 2D space without interrupting the gradient flow. 
This way, the scene is seamlessly stored in RAM or disk and can only be partially loaded and updated, which is crucial for scalability.

\boldparagraph{Appearance Rendering Module.}
In a first step, this module extracts compressed scene content $S_t$ using $\Pi$ and embeds these features into a higher dimensional space via a transposed linear compressor (\textit{Feature Decompression $W^T$}) which acts as a regularizer. 
Pre-computed meta features $\hat{M}_{t-1}$ are optionally appended and all features are downsampled by a custom masked average pooling with a stride of $k$ and $k \times k$ kernel size, where the mask indicates which features to ignore (features that are empty or located outside the scene space). 
The current $H \times W$ resolution feature map is passed through the seven-layer convolutional \textit{Rendering Network} refining features and filling potential holes that occur when the scene representation is sparsely populated. 
Lastly, the high-level features are decoded to RGB values by (\textit{Feature Decoder}) three linear layers interleaved with Leaky ReLU activation functions. 
The final output is activated using $\mathrm{HardTanh}$ activation for generating valid normalized RGB values.

\boldparagraph{Loss and Optimization.}
The entire pipeline is trained end-to-end from scratch until convergence using the reprojection error between the rendered image $\hat{I}_t$ and input image $I_t$ as self-supervision. 
Thus, the network can learn to optimally fuse and encode appearance information from 2D training data without any ground-truth textures.
Our pipeline is trained using a weighted combination of $\mathcal{L}_1$ and $\mathcal{L}_2$ loss between input image $I_t$ and rendered image $\hat{I}_t$ given by
\begin{equation} \label{eq:loss}
    \mathcal{L}(I_t, \hat{I}_t) = 
        \frac{1}{C\cdot H\cdot W} \!\!
            \sum_{p \in I_t, \hat{p} \in \hat{I}_t} \!\! ||p - \hat{p}||_1 + \frac{1}{2}||p - \hat{p}||_2
\end{equation}
We empirically found that a $1:\frac{1}{2}$ weight ratio worked best in our experiments.
The entire pipeline has less than 0.6M parameters and was optimized using the Adam optimizer~\cite{kingma2014adam} with a learning rate of $10^{-4}$ and batch size $1$, except for the generalization experiment, where we used 2.
Please see the supplementary material for more details.%

%% file: tex/5_experiments.tex
\begin{figure*}[tb]
    \vspace{-2pt}
    \centering
    \scriptsize
    \setlength{\tabcolsep}{2.6mm} %
    \newcommand{\sz}{0.094}  %
    \begin{tabular}{cccccccc}
           &    & Ours   &          & Texture & TSDF  & Waechter & Surfel- \\
        GT & Ours & Deterministic & Fu \etal~\cite{Fu-et-al-CVPR-2018} & Fields~\cite{Oechsle-et-al-ICCV-2019} & Coloring~\cite{seitz1999voxelcoloring} & \etal~\cite{Waechter-et-al-ECCV-14} & Meshing~\cite{Schoeps-et-al-TPAMI-2020} \\
        \includegraphics[width=\sz\textwidth,trim={195 73 195 65},clip]{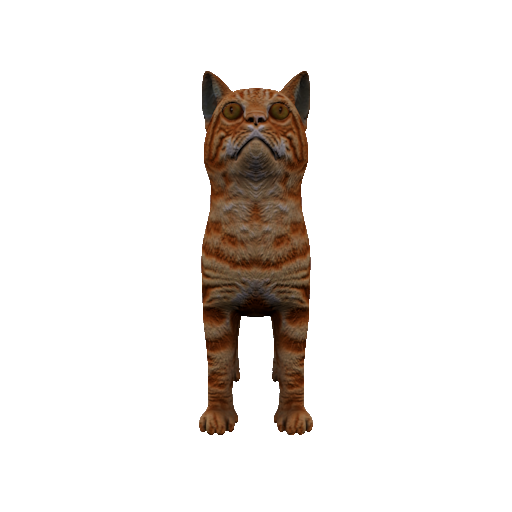} &
        \includegraphics[width=\sz\textwidth,trim={195 73 195 65},clip]{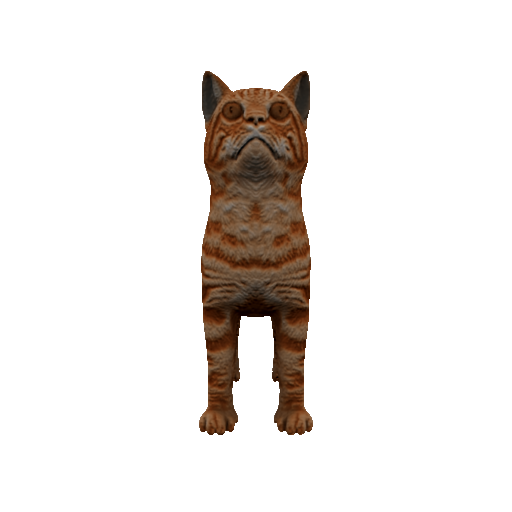} &
        \includegraphics[width=\sz\textwidth,trim={195 73 195 65},clip]{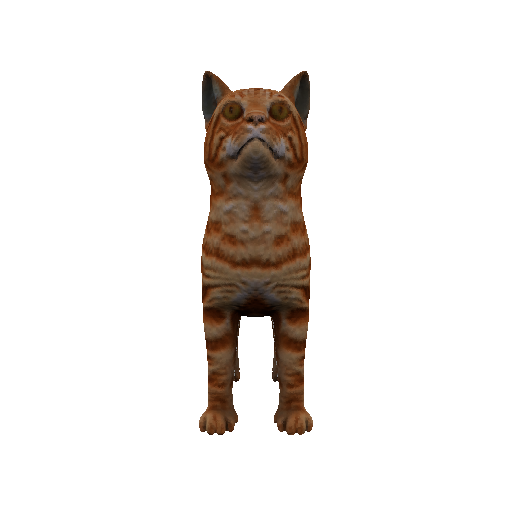} &
        \includegraphics[width=\sz\textwidth,trim={195 73 195 65},clip]{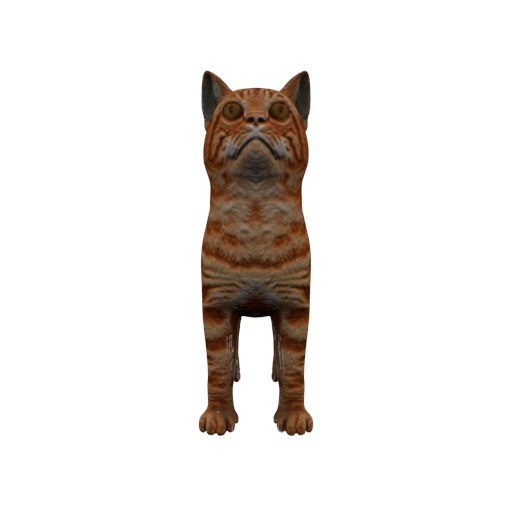} &
        \includegraphics[width=\sz\textwidth,trim={195 73 195 65},clip]{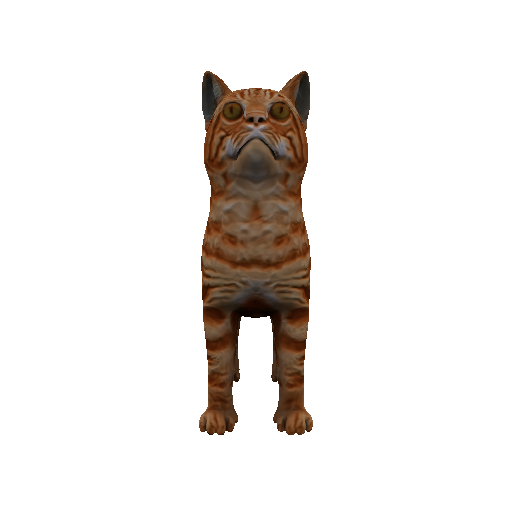} &
        \includegraphics[width=\sz\textwidth,trim={195 73 195 65},clip]{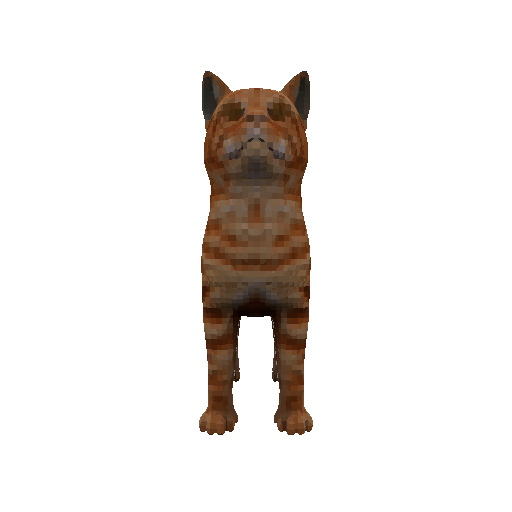}  &
        \includegraphics[width=\sz\textwidth,trim={195 73 195 65},clip]{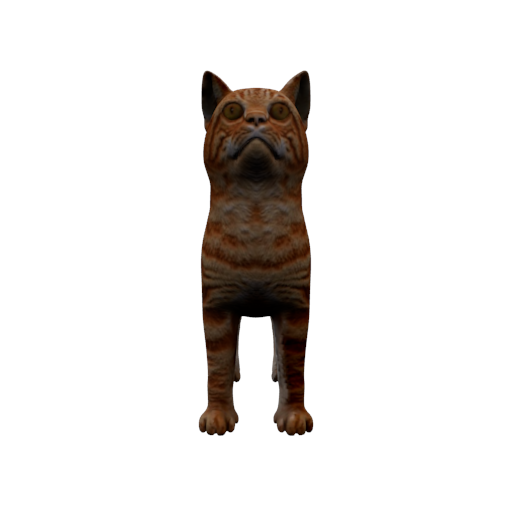} &
        \includegraphics[width=\sz\textwidth,trim={195 73 195 65},clip]{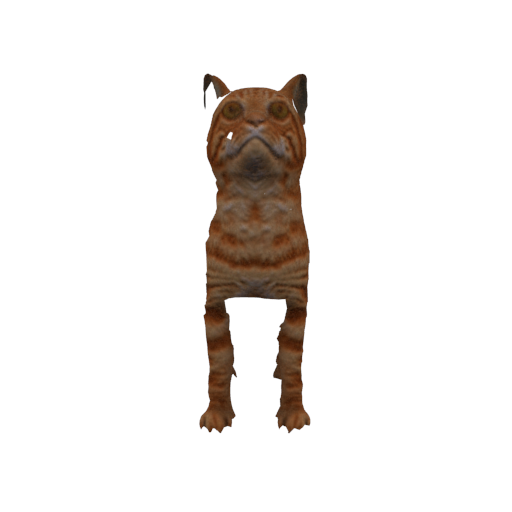} \\
        PSNR$\uparrow$ / SSIM$\uparrow$: & {\bf 32.94} / {\bf 0.950} & 29.44 / 0.889 & 32.14 / 0.912 & 27.99 / 0.856 & 24.89 / 0.621 & 18.52 / 0.631 & 9.928 / 0.510 \\[3pt]
        \includegraphics[width=\sz\textwidth,trim={190 30 180 5},clip]{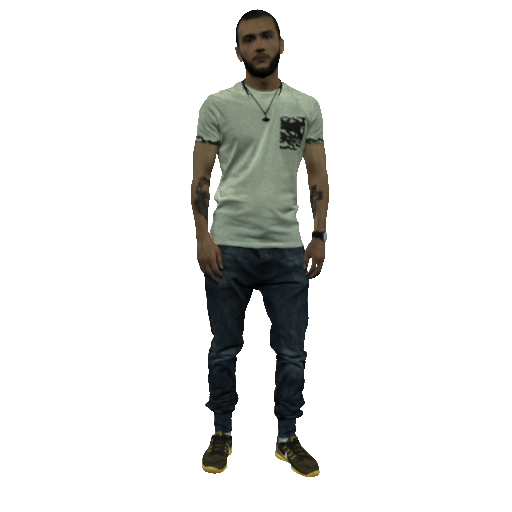} &
        \includegraphics[width=\sz\textwidth,trim={190 30 180 5},clip]{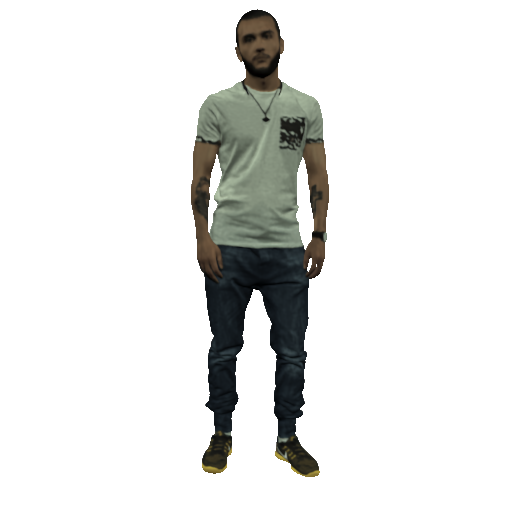} &
        \includegraphics[width=\sz\textwidth,trim={190 30 180 5},clip]{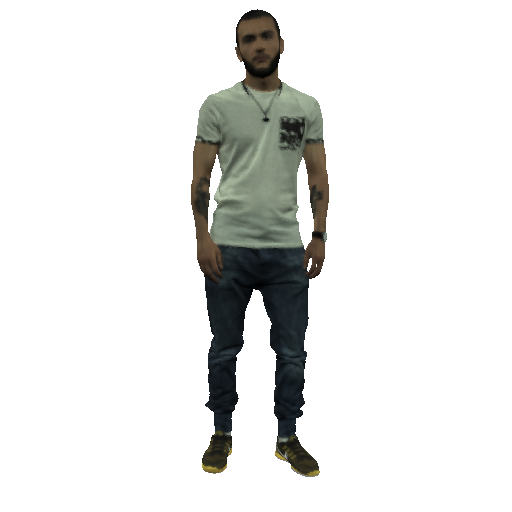} &
        \includegraphics[width=\sz\textwidth,trim={190 30 180 5},clip]{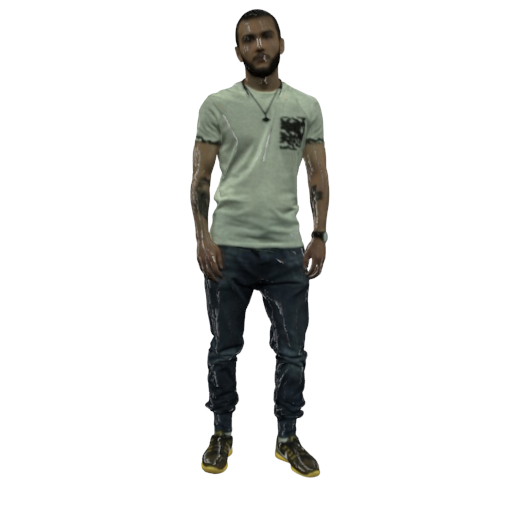} &
        \includegraphics[width=\sz\textwidth,trim={190 30 180 5},clip]{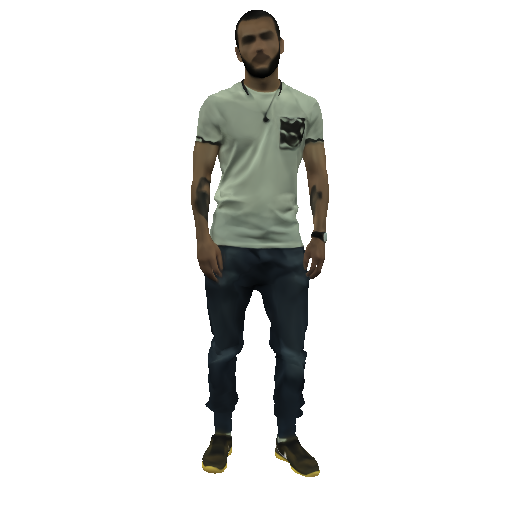} &
        \includegraphics[width=\sz\textwidth,trim={190 30 180 5},clip]{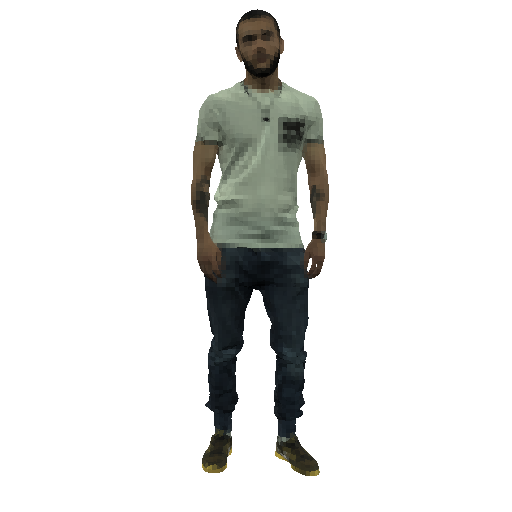} &
        \includegraphics[width=\sz\textwidth,trim={190 30 180 5},clip]{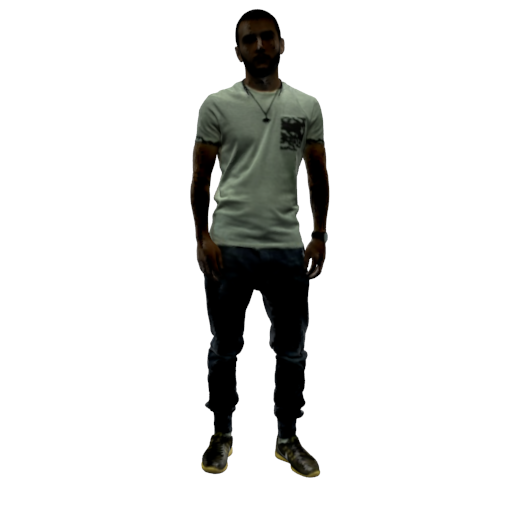} &
        \includegraphics[width=\sz\textwidth,trim={190 30 180 5},clip]{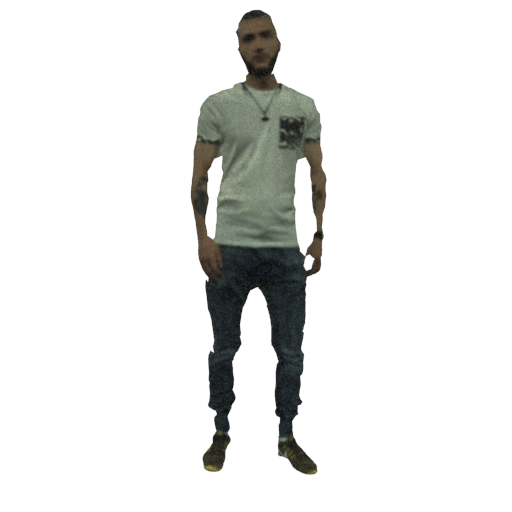} \\
        PSNR$\uparrow$ / SSIM$\uparrow$: & {\bf 35.42} / {\bf 0.966} & 32.03 / 0.958 & 27.23 / 0.835 & 27.80 / 0.841 & 27.37 / 0.810 & 16.24 / 0.370 & 12.31 / 0.458 \\[-2pt]
    \end{tabular}
    \caption{\textbf{Qualitative and quantitative comparison} on novel view synthesis with DeepSurfels on a $128^3$ sparse grid with learned 3-channel $4 \times 4$ feature patches. The experiment demonstrates that our scene representation is able to better represent high-frequency textures compared to other state-of-the-art methods. "Ours deterministic" shows direct rendering from RGB surfel patches. Please note that SurfelMeshing~\cite{Schoeps-et-al-TPAMI-2020} is the only method in this comparison which also estimates geometry while the other methods use known geometry. }
    \label{fig:nvs_cat_human}
\end{figure*}
\begin{figure*}[tb]
    \centering
    \includegraphics[width=\textwidth]{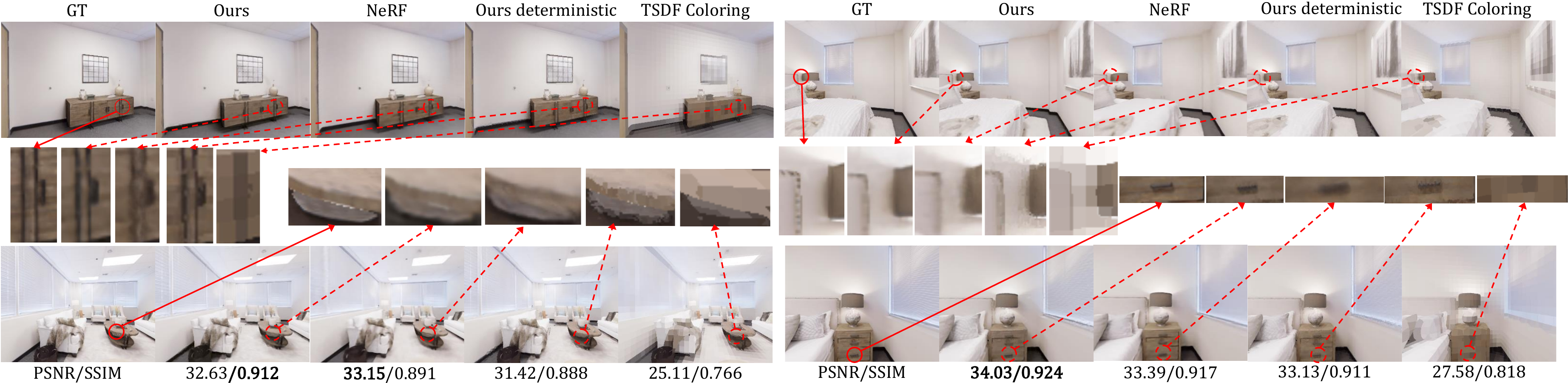}\\[-2pt]
    \caption{\textbf{Novel view synthesis for Replica~\cite{replica19arxiv} indoor scenes}. 
    The figure shows different views on two scenes (left and right). 
    Our learned approach has been trained only on the room on the left.
    NeRF~\cite{mildenhall2020nerf} is optimized separately on both scenes. 
    }
    \label{fig:nvs_replica}
\end{figure*}

\begin{figure*}[tb!]
    \centering
    \includegraphics[width=\textwidth]{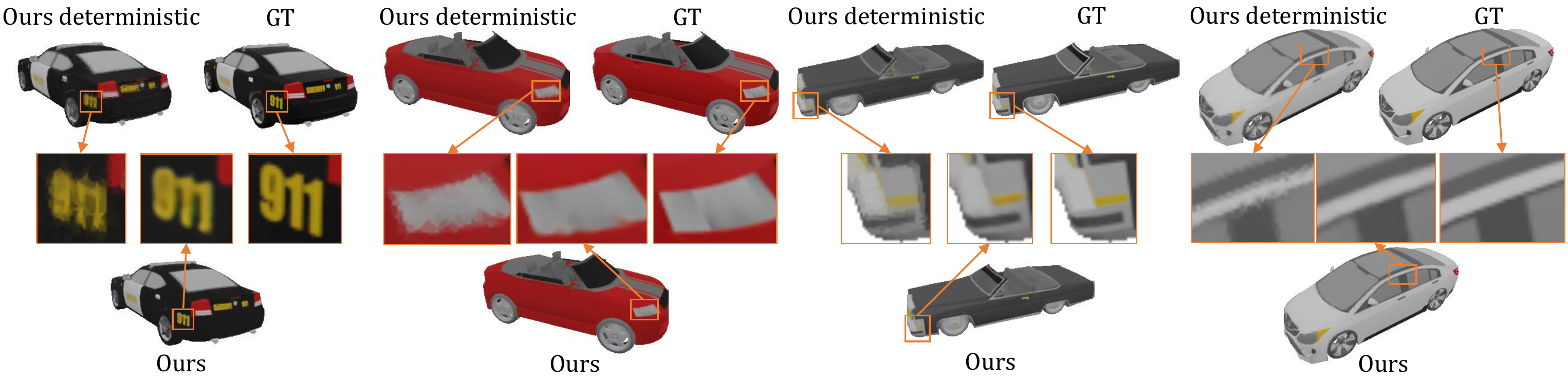}\\[-4pt] %
    \caption{\textbf{Qualitative results of our model on unseen scenes from ShapeNet~\cite{chang2015shapenet}.} Compared to deterministic rendering from RGB values, DeepSurfels with learned (3+3)-channel $6\!\times\!6$ patches on a sparse $32^3$ grid yields significantly more high-frequency details. As it is shown in the close ups, storing learned features in the texels is particularly useful to correct discretization artefacts.}
    \label{fig:generalization}
\end{figure*}
\begin{figure}[tb]
    \centering
    \scriptsize
    \setlength{\tabcolsep}{0.4mm}
    \newcommand{\sza}{0.16}
    \newcommand{\szb}{0.16}
    \begin{tabular}{cccccc}
        GT & Ours    & SRNs~\cite{Sitzmann-et-al-NIPS-2019}   & GT        & Ours   & DeepVoxels \\ %
        \hspace{-5pt}
        \includegraphics[width=\sza\textwidth,trim={120 110 108 120},clip]{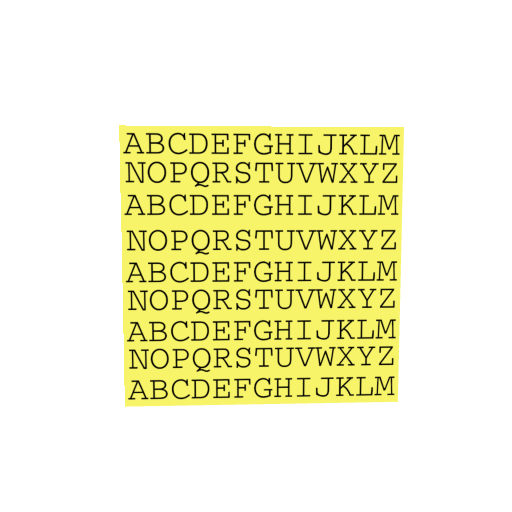} &
        \includegraphics[width=\sza\textwidth,trim={120 110 108 120},clip]{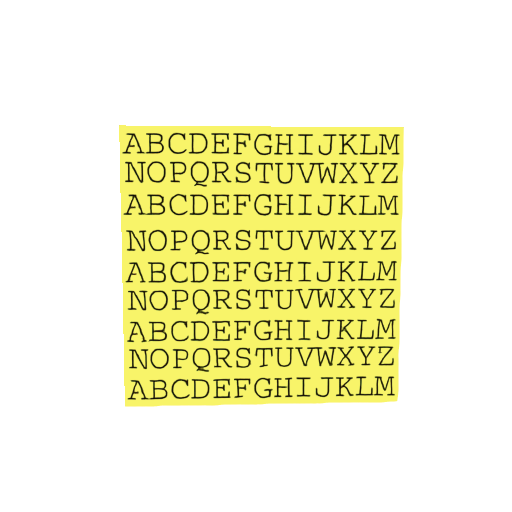} &
        \includegraphics[width=\sza\textwidth,trim={120 110 108 120},clip]{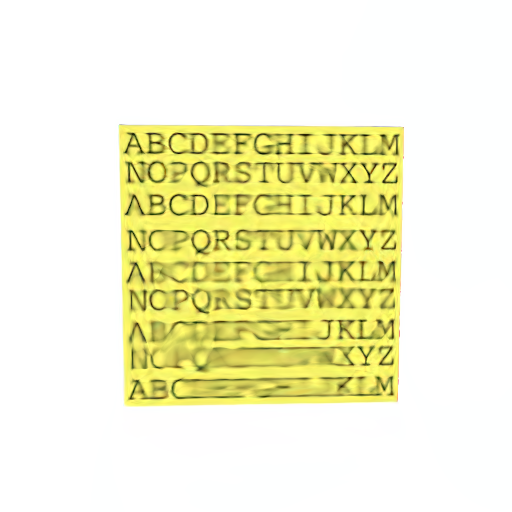} &
        \includegraphics[width=\szb\textwidth,trim={91 102 120 102},clip]{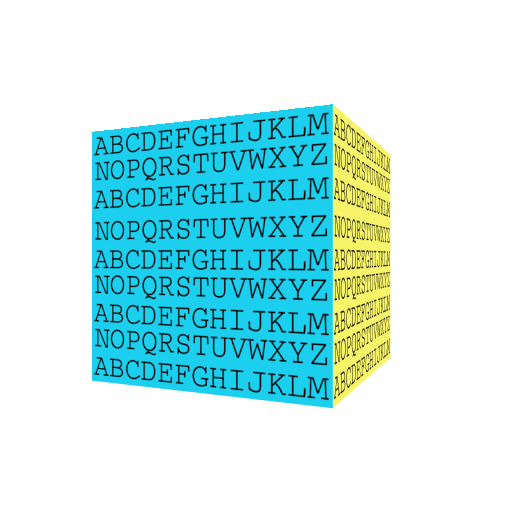} &
        \includegraphics[width=\szb\textwidth,trim={91 102 120 102},clip]{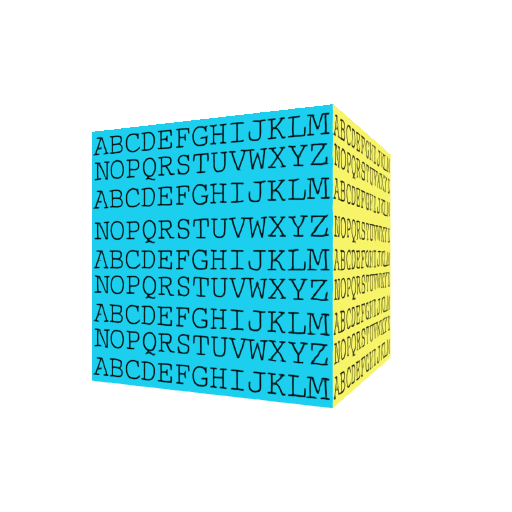} &
        \includegraphics[width=\szb\textwidth,trim={91 102 120 102},clip]{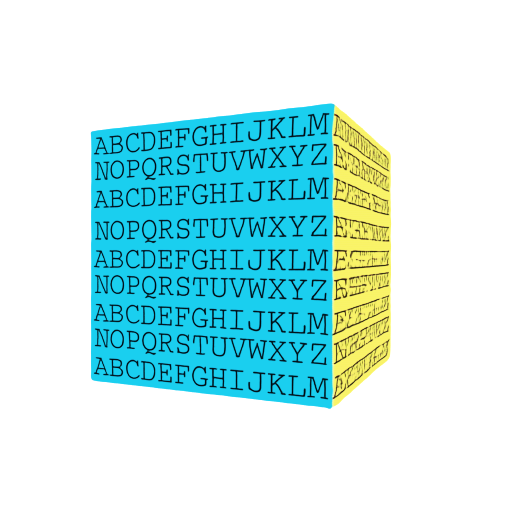} \\
        \hspace{-2pt}\tiny PSNR$\uparrow$ / SSIM$\uparrow$: & {\bf 22.19} / {\bf 0.93}  & 16.74 / 0.65 & & {\bf 22.54} / {\bf 0.91} & 20.12 / 0.84 \\%[3pt] New results for our 64^1x1 method with 8 channels
    \end{tabular}
    \vspace{-4pt}
    \caption{\textbf{Comparison of SRNs~\cite{Sitzmann-et-al-NIPS-2019} and DeepVoxels~\cite{Sitzmann-et-al-CVPR-2019}} to our learned DeepSurfel fusion with a $64^3$  grid of 8-channel $1 \times 1$ resolution feature patches on the synthetic cube dataset from \cite{Sitzmann-et-al-CVPR-2019}. Our method produces fewer blur artifacts and multi-view inconsistencies and overall yields significantly better images reconstruction results than both baselines.
    Note that both baselines perform global appearance fusion with unknown geometry.}
    \label{fig:nvs_sitzmann}
\end{figure}

\section{Evaluation}
We evaluate our method by comparing the representation power of both, our learned and our deterministic approach (direct rendering from RGB surfels) with state-of-the-art methods on novel view synthesis tasks. 
We further demonstrate how our method generalizes to different scenes for a small number of distinct training samples and provide an ablation study to validate the design choices for our model. 
The supplementary material provides further details.

\boldparagraph{Datasets.}
We conduct experiments on datasets generated from Shapenet~\cite{chang2015shapenet}, publicly available human and cat\footnote{3D models from \url{free3d.com} and \url{turbosquid.com}.} models, the indoor Replica dataset~\cite{replica19arxiv}, and the cube scene from~\cite{Sitzmann-et-al-CVPR-2019}.
Replica dataset images were rendered with Habitat-Sim~\cite{habitat19iccv} and all other models with Blender~\cite{blender20}.

\boldparagraph{Metrics.}
We quantify model performances with the following two metrics~\cite{wang2004image}.  
\textbf{PSNR:} The Peak Signal-to-Noise Ratio is the ratio between the maximum pixel value in the ground-truth image and the pixel-wise mean-squared error between ground-truth and rendered image.
\textbf{SSIM:} The Structural Similarity Index measures similarity between patches of rendered and ground-truth images. 
We omit other perceptron based metrics because we are interested in recovering the true pixel value as our fusion approach can be used for more general types of data. 

\boldparagraph{Novel View Synthesis.} 
The model is optimized on $500$ randomly rendered $512 \times 512$ training images for the cat and human model and the results for a single unseen frontal viewpoint\footnote{For a fair comparison with the results of Texture Fields~\cite{Oechsle-et-al-ICCV-2019}.} are compared with state-of-the-art batch (Fu \etal~\cite{Fu-et-al-CVPR-2018}, Texture Fields~\cite{Oechsle-et-al-ICCV-2019}, Waechter \etal~\cite{Waechter-et-al-ECCV-14}), and online methods (SurfelMeshing~\cite{Schoeps-et-al-TPAMI-2020}, TSDF Coloring~\cite{curless1996volumetric}) on a $128^3$ grid. 
The results in \figurename~\ref{fig:nvs_cat_human} demonstrate that our approach compares favorably even to slower batch-based methods in representing high-frequency textures.
\figurename~\ref{fig:nvs_sitzmann} further shows that our approach does not suffer from blurry artifacts as the recently proposed SRNs~\cite{Sitzmann-et-al-NIPS-2019}, or from multi-view consistency issues like DeepVoxels~\cite{Sitzmann-et-al-CVPR-2019}. 
Note that these approaches jointly estimate geometry and appearance while we only estimate appearance. 
\tablename~\ref{tab:channels_study} shows the effect of varying the number of channels on the cube dataset for DeepSurfels of $4 \times 4$ patches on a $64^3$ sparse grid.

\boldparagraph{Generalization.} 
Our pipeline scales and generalizes well on realistic room-size scenes. 
We trained our pipeline on $288$ $480 \times 640$ images of one Replica~\cite{replica19arxiv} room represented with DeepSurfels of $11$cm voxel size with (3+3)-channel $6 \times 6$ resolution patches. 
We disentangled 3 color channels to improve generalization. 
The pipeline evaluation is performed on every 25th unseen frame in a sequence of frames generated by a moving agent in the Habitat Sim~\cite{habitat19iccv}.
\figurename~\ref{fig:nvs_replica} shows results of our trained pipeline on an optimized (left) and a non-optimized (right) scene. 
Our learned approach outperforms baselines in representing fine details.
\begin{table}[tb]
    \centering
    \scriptsize
    \setlength{\tabcolsep}{0.5pt}
    \renewcommand{\arraystretch}{1.27}
    \newcommand{\ccol}[1]{\textcolor{CadetBlue}{#1}} %
    \floatbox[\capbeside]{table}[0.54\columnwidth]%
    {\caption{\textbf{Ablation study} on ShapeNet~\cite{chang2015shapenet} cars. The \textbf{top part} of the table compares various baselines. Our deterministic coloring at $32^3$ is still better than TSDF Coloring at $128^3$ resolution. The \textbf{mid part} shows the impact of the proposed losses. The \textbf{bottom part} shows the influence of the voxel grid, surfel patch and channel resolution, demonstrating that quality improvements saturate for higher resolutions. 3+3 denotes 3 feature and 3 color channels (disentangled) per texel. We also compare to \ccol{$1\!\!\times\!\!1$ patches with 213+3 channels} corresponding to the same number of features as for $6\!\!\times\!\!6$, 3+3, demonstrating the benefit of a spatial sub-feature alignment in our network. } 
    \label{tab:ablation_study}}
    {\begin{tabular}[b]{clcc}
      \toprule
      & Method & PSNR$\uparrow$ & SSIM$\uparrow$ \\ 
      \midrule
      \multirow{8}{*}{\rotatebox{90}{Baselines}}
      & SurfelMeshing~\cite{Schoeps-et-al-TPAMI-2020}           & 13.92 & 0.2748 \\
      & Waechter \etal~\cite{Waechter-et-al-ECCV-14}            & 18.27 & 0.4753 \\
      & Fu \etal~\cite{Fu-et-al-CVPR-2018}                      & 18.84 & 0.5196 \\ %
      & TSDF Coloring~\cite{curless1996volumetric} ($32^3$)     & 21.57 & 0.6375 \\
      & TSDF Coloring~\cite{curless1996volumetric} ($64^3$)     & 24.05 & 0.7552 \\
      & TSDF Coloring~\cite{curless1996volumetric} ($128^3$)    & 26.68 & 0.8526 \\
      & Ours Det. ($32^3$, $6 \!\!\times\!\! 6$, 3)             & 27.20 & 0.8723 \\
      & Ours Det. ($64^3$, $4 \!\!\times\!\! 4$, 3)             & 28.73 & 0.9036 \\[0.8pt] \hdashline \noalign{\vskip 1pt}
      \multirow{5}{*}{\rotatebox{90}{Ablation}}
      & Learned ($32^3$, $6 \!\!\times\!\! 6$, 3+3)             & 28.27 & 0.8777 \\
      & + depth                                                 & 28.31 & 0.8782 \\
      & + multi-view consist.                                   & 28.36 & 0.8889 \\
      & + viewing direction \& \\
      & \phantom{+} surface orientation                         & {\bf 28.89} & {\bf 0.8907} \\[0.8pt] \hdashline \noalign{\vskip 1pt} 
    \multirow{9}{*}{\rotatebox{90}{DeepSurfel Parameters}}
      & \phantom{$1$}\ccol{$32^3$, $1 \!\!\times\!\! 1$, 213+3} & \ccol{22.95} & \ccol{0.7083} \\
      & \phantom{$1$}\ccol{$64^3$, $1 \!\!\times\!\! 1$, 213+3} & \ccol{25.41} & \ccol{0.7940} \\
      & \phantom{$1$}$64^3$, $4 \!\!\times\!\! 4$, 3+3          & 29.92 & 0.9086 \\ %
      & \phantom{$1$}$64^3$, $5 \!\!\times\!\! 5$, 3+3          & 30.15 & 0.9126 \\
      & \phantom{$1$}$64^3$, $6 \!\!\times\!\! 6$, 3+3          & {\bf 30.27} & {\bf 0.9147} \\[2pt]
      & \ccol{$128^3$, $1 \!\!\times\!\! 1$, 213+3}             & \ccol{26.75} & \ccol{0.8324} \\
      & $128^3$, $2 \!\!\times\!\! 2$, 3+3                      & 30.23 & 0.9133 \\
      & $128^3$, $3 \!\!\times\!\! 3$, 3+3                      & 30.51 & 0.9181 \\
      & $128^3$, $4 \!\!\times\!\! 4$, 3+3                      & 30.60 & 0.9196 \\
      & $128^3$, $5 \!\!\times\!\! 5$, 3+3                      & 30.63 & 0.9200 \\
      & $128^3$, $6 \!\!\times\!\! 6$, 3+3                      & {\bf 30.64} & {\bf 0.9202} \\
      \bottomrule
    \end{tabular}}
    \vspace{-1em}
\end{table}

We further demonstrate that our pipeline generalizes well when trained on a larger set of distinct scenes. 
We render 100 $312 \times 312$ training images from 150 Shapenet~\cite{chang2015shapenet} car scenes and test the pipeline on 50 unseen scenes by fusing 100 views and evaluating results on additional 60 unseen viewpoints.
The whole pipeline is trained to be frame order independent by randomly shuffling scenes and frames after each optimization step. 
Results on test scenes (\figurename~\ref{fig:generalization}, \tablename~\ref{tab:ablation_study}) indicate that our learned approach improves for discretization artifacts and overall yields sharper results which are supported by higher PSNR and SSIM scores.

\boldparagraph{Ablation Study.}
For the unobserved test car scenes, we quantify in \tablename~\ref{tab:ablation_study} the impact of: 
\textbf{(i)} depth as a meta feature that helps our method to reason about the confidence of updates since pixels with larger depth values are less important; %
\textbf{(ii)} multi-view consistency regularization that corrects for geometric misalignments and improves interpolation among neighboring viewpoints by adding an additional error signal (\ref{eq:loss}) for a viewpoint closest to the fused frame; 
\textbf{(iii)} pixel ray directions with surface orientation map to improve reasoning about light information and non-Lambertian surfaces; 
\textbf{(iv)} DeepSurfel parameters (grid and patch resolution). For almost all experiments use 3 feature and 3 disentangled color channels (denoted as 3+3 in \tablename~\ref{tab:ablation_study}) which outperforms the baselines. 
We observe that every attribute (i-iii) improves generalization and that higher DeepSurfel resolution (iv) consistently benefits reconstruction quality.
\textbf{(v)} Lastly, we demonstrate the benefit of explicitly modeling the reprojection of pixel colors via texels which can be seen as subfeatures of a large feature vector stored in every voxel.
We compare the proposed $6 \times 6$ patches, 3+3 channels to $1 \times 1$ patches, 213+3 channels which amounts to the same number of features per voxel.
We argue that the benefit of using more features per voxel quickly saturates unless subfeatures are anchored along the surface and trained with separate pixel data as supported by our results.
\begin{table}[t]%
    \centering
    \scriptsize
    \setlength{\tabcolsep}{7pt}
	\renewcommand{\arraystretch}{1.3}
    \floatbox[\capbeside]{table}%
    {\caption{\textbf{Varying number of feature channels} for the cube~\cite{Sitzmann-et-al-CVPR-2019} dataset on $64^3$ sparse grid with $4 \times 4$ patches. Additional feature channels improve the reconstruction quality.}
    \label{tab:channels_study}}%
    {\begin{tabular}{lcc}
      \toprule
      \#Channels    & PSNR$\uparrow$  & SSIM$\uparrow$ \\ 
        \midrule
        2	& 25.95	& 0.9432 \\
        4	& 26.72	& 0.9506 \\
        6	& 27.33	& 0.9568 \\
        10	& {\bf 28.27}	& {\bf 0.9638} \\
      \bottomrule
    \end{tabular}
    }
    \vspace{-5pt}
\end{table}

\begin{figure}[t]
   \centering
   \includegraphics[width=\linewidth]{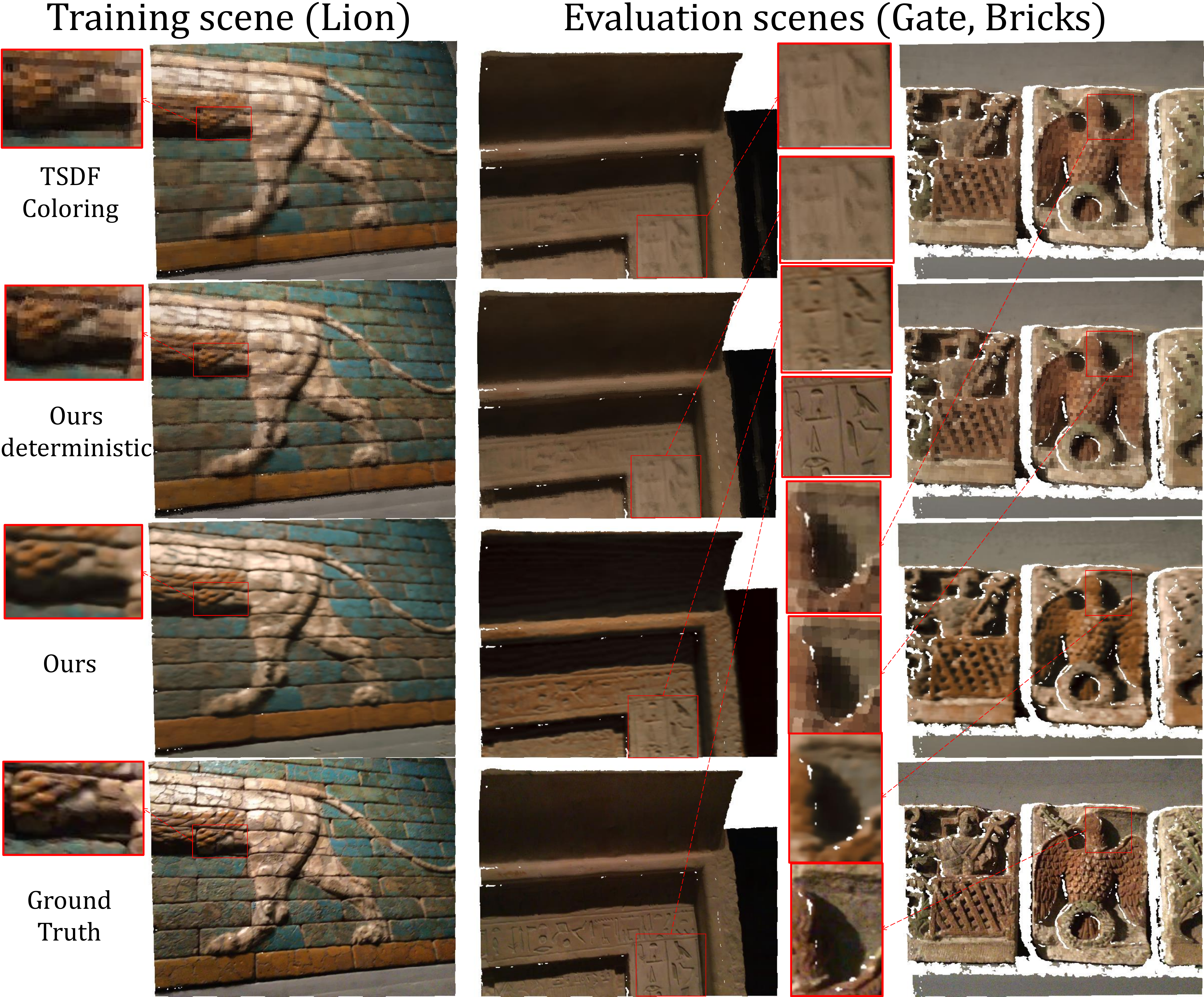}
   \vspace{-10pt}
   \caption{\textbf{Novel-view synthesis on unseen real-world data~\cite{Maier-et-al-ICCV-2017}.}
            Our DeepSurfel method with $4\times4$ patches is trained on the \emph{Lion} scene for 80 training iterations and then evaluated on the \emph{Gate} and the \emph{Bricks} scenes. 
            The images show novel viewpoints.
            }
\label{fig:real_data}
\vspace{-5pt}
\end{figure}
\boldparagraph{Real-world data.}
In Fig.~\ref{fig:real_data} we present results on unseen real-world data from~\cite{Maier-et-al-ICCV-2017} for which our method yields the most detailed appearance reconstructions.

\boldparagraph{Runtime.}
Our method takes $57$ms and $21$ms for fusing and rendering a single $312 \times 312$ frame on $32^3$ DeepSurfels with 6-channel patches with resolution $6 \times 6$ (\tablename~\ref{tab:ablation_study}).
This is significantly faster compared to other deep learning methods that overfit on a single scene. 
For example, the state-of-the-art method NeRF~\cite{mildenhall2020nerf} requires $\sim2$ days for training on a single scene being unable to generalize to other scenes, while our method can easily be used on unseen scenes without any optimization as demonstrated in \figurename~\ref{fig:nvs_replica}, which is a speed up of over a thousand times on unobserved scenes for comparable or even favorable results.

\boldparagraph{Limitations.}
Similar to traditional methods like TSDF Coloring, our method is sensitive to camera-geometry misalignment which can lead to blurry results. %
Moreover, we currently do not model view-dependent effects which may result in washed-out colors due to the local feature averaging.
When trained on small datasets our method may distort colors that were not seen during training.

%% file: tex/6_conclusion.tex
\section{Conclusion}
We introduced DeepSurfels, a novel scene representation for geometry and appearance encoding, that combines explicit and implicit scene representation to improve for scalability and interpretability. 
It is defined on a sparse voxel grid to maintain topology relations and implements 2D geometry oriented patches to store high-frequency appearance information.
We further presented a learned approach for online appearance fusion that compares favorably to existing offline and online texture mapping methods since it learns to correct for typical noise and discretization artifacts. %

As future work we consider the joint online fusion of shape and appearance and address some weaknesses of our appearance fusion pipeline such as the limitation in filling large missing parts and rendering translucent surfaces.

%% file: tex/8_acknowledgements.tex
This research was partly supported by Toshiba and Innosuisse funding (Grant No. 34475.1 IP-ICT).

%% file: supplementary_material/main.tex
\clearpage
\begingroup

\twocolumn[
\begin{center}
  {\Large \bf \Large{\bf {DeepSurfels}: Learning Online Appearance Fusion} \\ -- Supplementary Material -- \par}
  \vspace*{12pt}
\end{center}
]
\appendix

\setcounter{page}{1}
\setcounter{table}{0}
\setcounter{figure}{0}
\setcounter{equation}{0}
\renewcommand{\thetable}{\thesection.\arabic{table}}
\renewcommand{\thefigure}{\thesection.\arabic{figure}}
\renewcommand{\theequation}{\thesection.\arabic{equation}}

\section{Overview}
In this supplementary document, we provide further details about our appearance learning pipeline (\S~\ref{reproducibility}), used baselines (\S~\ref{baselines}), and an extended ablation study (\S~\ref{ext_ablation}).
\input{supplementary_material/tex/reproducibility}

\input{supplementary_material/tex/ablation_study}

%% file: supplementary_material/tex/reproducibility.tex
\section{Network architecture details} \label{reproducibility}
We provide further details on the Fusion Network and the Appearance Rendering module presented in \figurename~\ref{fig:full_pipeline}.

\begin{figure*}
    \centering
    \includegraphics[width=0.7\textwidth]{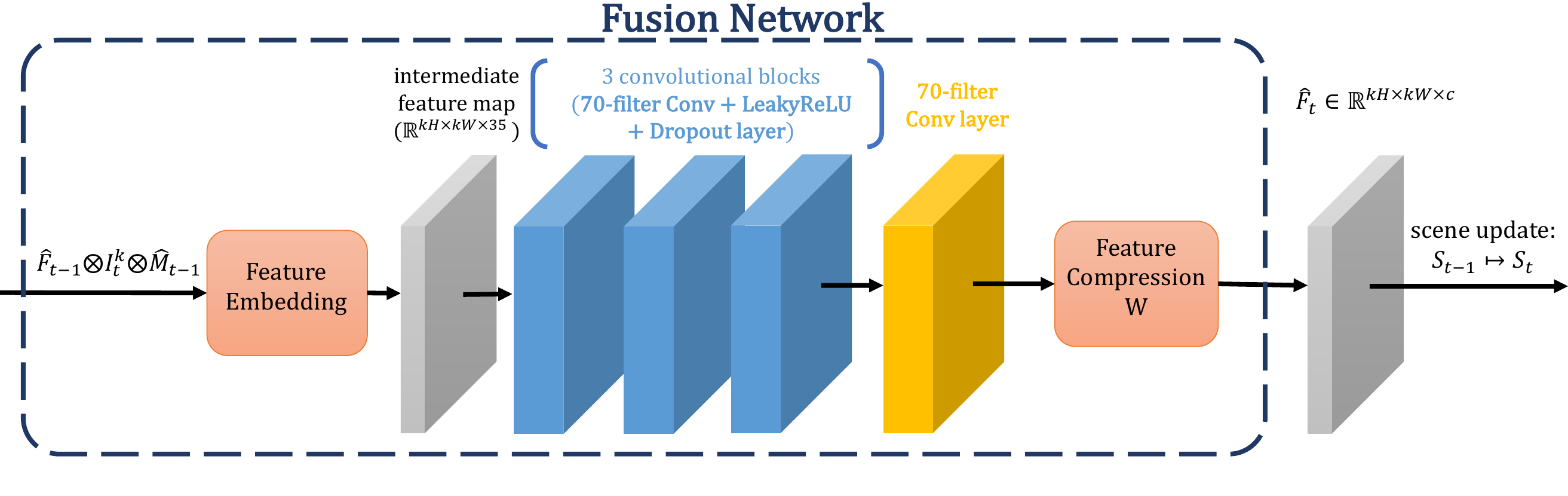} \\[-10pt]
    \caption{\textbf{Fusion network architecture.}
    This module is a part of our learned appearance fusion pipeline (\figurename~\ref{fig:full_pipeline}). 
    It creates a blended feature map $\hat{F}_{t}$ that needs to be integrated into DeepSurfel representation..
    }
    \label{fig:app:fusion_network}
\end{figure*}

The Fusion Network is displayed in \figurename~\ref{fig:app:fusion_network} and represents one part of the Appearance Fusion module (\figurename~\ref{fig:full_pipeline}).
It takes as input three image maps -- the upsampled input image that needs to be fused $I_t^k \in \mathbb{R}^{k H \times k W \times C}$, a feature map $\hat{F}_{t-1}$ that is rendered from existing scene content $S_{t-1}$, and optional meta features $\hat{M}_{t-1}$ -- and produces a new blended feature map $\hat{F}_{t}$ that needs to be integrated into the representation.

This component consists of three modules
\emph{1)} the Feature Embedding learnable linear layer, implemented as a $1 \times 1$ convolutional layer, which compresses features of the concatenated input maps ($\hat{F}_{t-1} \otimes \hat{M}_{t-1} \otimes I_t^k$) into an intermediate feature map $\mathbb{R}^{kH \times kW \times 35}$,
\emph{2)} the Blending Network that comprises of four convolutional blocks interleaved with LeakyReLU and dropout layers, and
\emph{3)} the linear Feature Compression layer $W: \mathbb{R}^{kH \times kW \times 70} \mapsto \mathbb{R}^{kH \times kW \times c}$ that creates the new blended feature map $\hat{F}_{t}$.
This new feature map is then integrated into the scene representation as described in the paper by updating the scene content ($S_{t-1} \mapsto S_t$).

\begin{figure*}
    \centering
    \includegraphics[width=\textwidth]{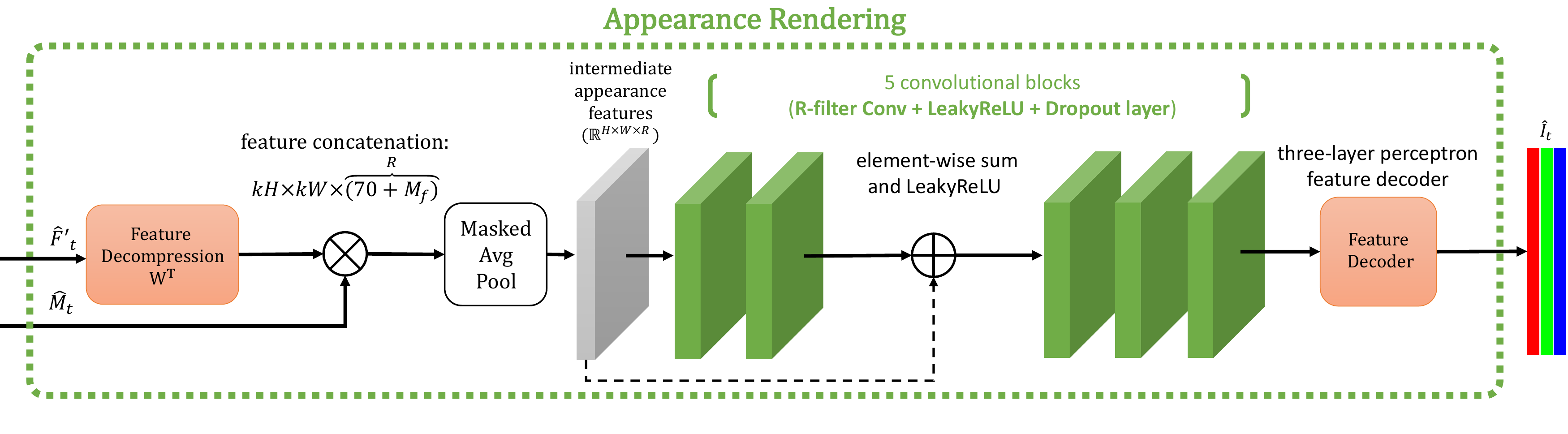} \\[-10pt]
    \caption{\textbf{Appearance rendering module.}
    This module interprets rendered feature as RGB pixel values.
    $M_f$ denotes the number of feature channels and $R$ is the number of channels of the intermediate appearance features ($R=70+M_f$).
    The intermediate appearance features are refined by 5 convolutional blocks and decoded by a Feature Decoder.
    The Feature Decoder is implemented as a three-layer perceptron network with $\floor{\frac{R}{2}}$, $\floor{\frac{R}{4}}$, and $3$ neurons respectively, each layer is followed by LeakyReLU activation function, except for the very last one that uses HardTanh to produce normalized RGB color values.
    }
    \label{fig:app:appearance_rendering_module}
\end{figure*}

The updated scene content is then rendered $\hat{F}_{t}^\prime \in \mathbb{R}^{k H \times k W \times c}$ via the introduced differentiable projection module $\Pi$.
The Appearance Rendering module (\figurename~\ref{fig:app:appearance_rendering_module}) takes this rendered feature map and decompresses its features into a higher resolution space with the linear Feature Decompression layer (transposed Feature Compression layer $W^T: \mathbb{R}^{kH \times kW \times c} \mapsto \mathbb{R}^{kH \times kW \times 70}$).
The optional meta features are concatenated to the uncompressed feature channels and they are jointly propagated through the introduced masked average pooling operator to reduce the spatial dimension ($kH, kW \mapsto H, W$) and form an intermediate appearance feature map.
This appearance feature map is then refined by the Rendering Network (5 convolutional blocks with a skip connection) and decoded as RGB values by the three-layer perceptron Feature Decoder.

\section{Baseline experiments} \label{baselines}
Several baselines are used in the paper for results displayed in \figurename~\ref{fig:nvs_cat_human},  \ref{fig:nvs_replica}, and \ref{fig:nvs_sitzmann}.

We used publicly released code with default parameters to run experiments for Fu \etal~\cite{Fu-et-al-CVPR-2018}\footnote{https://github.com/fdp0525/G2LTex}, SurfelMeshing~\cite{Schoeps-et-al-TPAMI-2020}\footnote{https://github.com/puzzlepaint/surfelmeshing},
Waechter \etal~\cite{Waechter-et-al-ECCV-14}\footnote{https://www.gcc.tu-darmstadt.de/home/proj/texrecon/}, and
NeRF~\cite{mildenhall2020nerf}\footnote{https://github.com/bmild/nerf}.
The results for other baselines (Texture Fields~\cite{Oechsle-et-al-ICCV-2019}, SRNs~\cite{Sitzmann-et-al-NIPS-2019}, DeepVoxels~\cite{Sitzmann-et-al-CVPR-2019}) are released by the authors and 
we implemented the TSDF Coloring~\cite{curless1996volumetric} baseline as a straightforward extension of TSDF Fusion that accumulates color information into voxel grids by the simple running mean algorithm.

Mesh files for Fu \etal~\cite{Fu-et-al-CVPR-2018} and Waechter \etal~\cite{Waechter-et-al-ECCV-14} for the experiment on the ShapeNet~\cite{chang2015shapenet} cars (\tablename~\ref{tab:ablation_study}) are created by fusing depth frames into a grid with TSDF Fusion and then extracting the meshes with a standard marching cubes algorithm. 
These methods where provided by the ground truth meshes for the novel view synthesis experiment on the cat and the human dataset (\figurename~\ref{fig:nvs_cat_human}).

NeRF~\cite{mildenhall2020nerf} was trained for each Replica room dataset (\figurename~\ref{fig:nvs_replica}) for two days on a 24GB NVidia Titan RTX GPU.

%% file: supplementary_material/tex/ablation_study.tex
\section{Ablation study} \label{ext_ablation}

\begin{table}
    \centering
    \setlength{\tabcolsep}{8pt}
    \begin{tabular}[b]{clcc}
      \toprule
      & Method & PSNR$\uparrow$ & SSIM$\uparrow$ \\ \midrule
      \multirow{8}{*}{\rotatebox{90}{Baselines}}
      & SurfelMeshing~\cite{Schoeps-et-al-TPAMI-2020}  & 13.92 & 0.2748  \\
      & Waechter \etal~\cite{Waechter-et-al-ECCV-14}  & 18.27 & 0.4753  \\
      & Fu \etal~\cite{Fu-et-al-CVPR-2018}            & 18.84 & 0.5196  \\ %
      & TSDF Coloring~\cite{curless1996volumetric} ($32^3$)   & 21.57 & 0.6375  \\
      & TSDF Coloring~\cite{curless1996volumetric} ($64^3$)   & 24.05 & 0.7552  \\
      & TSDF Coloring~\cite{curless1996volumetric} ($128^3$)  & 26.68 & 0.8526  \\
      & Ours Det. ($32^3$, $6 \!\!\times\!\! 6$, 3)    & 27.20 & 0.8723  \\
      & Ours Det. ($64^3$, $4 \!\!\times\!\! 4$, 3)    & 28.73 & 0.9036  \\[1.8pt] \hdashline \noalign{\vskip 3pt}

    \multirow{9}{*}{\rotatebox{90}{DeepSurfel Params (3+3)}}
      & \phantom{$1$}$32^3$, $6 \!\!\times\!\! 6, 3+3$ & {\bf 28.89} & {\bf 0.8907} \\[2pt]

      & \phantom{$1$}$64^3$, $4 \!\!\times\!\! 4, 3+3$ & 29.92 & 0.9086 \\
      & \phantom{$1$}$64^3$, $5 \!\!\times\!\! 5, 3+3$ & 30.15 & 0.9126 \\
      & \phantom{$1$}$64^3$, $6 \!\!\times\!\! 6, 3+3$ & {\bf 30.27} & {\bf 0.9147} \\[2pt]

      & $128^3$, $2 \!\!\times\!\! 2, 3+3$             & 30.23 & 0.9133 \\
      & $128^3$, $3 \!\!\times\!\! 3, 3+3$             & 30.51 & 0.9181 \\
      & $128^3$, $4 \!\!\times\!\! 4, 3+3$             & 30.60 & 0.9196 \\
      & $128^3$, $5 \!\!\times\!\! 5, 3+3$             & 30.63 & 0.9200 \\
      & $128^3$, $6 \!\!\times\!\! 6, 3+3$             & {\bf 30.64} & {\bf 0.9202} \\[1.8pt] \hdashline \noalign{\vskip 3pt}

    \multirow{9}{*}{\rotatebox{90}{DeepSurfel Params (5+3)}}
      & \phantom{$1$}$32^3$, $6 \!\!\times\!\! 6, 5+3$ & {\bf 29.02} & {\bf 0.8955} \\[2pt]

      & \phantom{$1$}$64^3$, $4 \!\!\times\!\! 4, 5+3$ & 29.93 & 0.9118 \\
      & \phantom{$1$}$64^3$, $5 \!\!\times\!\! 5, 5+3$ & 30.12 & 0.9154 \\
      & \phantom{$1$}$64^3$, $6 \!\!\times\!\! 6, 5+3$ & {\bf 30.22} & {\bf 0.9172} \\[2pt]

      & $128^3$, $2 \!\!\times\!\! 2, 5+3$             & 30.21 & 0.9162 \\
      & $128^3$, $3 \!\!\times\!\! 3, 5+3$             & 30.45 & 0.9206 \\
      & $128^3$, $4 \!\!\times\!\! 4, 5+3$             & 30.54 & 0.9220 \\
      & $128^3$, $5 \!\!\times\!\! 5, 5+3$             & 30.56 & 0.9224 \\
      & $128^3$, $6 \!\!\times\!\! 6, 5+3$             & {\bf 30.58} & {\bf 0.9226} \\
      \bottomrule\\%[2pt]
    \end{tabular}
    \vspace{-20pt}
    \caption{\textbf{Extended ablation study on ShapeNet~\cite{chang2015shapenet} cars.}
    Comparison of baselines and our method on different grid and patch resolutions. The x+3 notation denotes disentangled x feature channels and 3 color channels.
    Results indicate that our method on a grid of $32^3$ outperforms all baseline methods, including ones that require a much higher grid resolution (TSDF Coloring $128^3$, Ours Deterministic $64^3$).
    An increased number of channels and higher DeepSurfel resolution further benefits the quality of rendered images.
    Qualitative results for the 3+3 and 5+3 configuration are displayed in \figurename~\ref{fig:app:nvs_ch3} and \ref{fig:app:nvs_ch5} respectively.
    }
    \label{tab:app:ablation_study}
\end{table}

We provide an extended ablation study for 5 feature and 3 color channels (5+3 configuration) in comparison to the 3+3 configuration in \tablename~\ref{tab:app:ablation_study}.

\begin{figure*}
    \centering
    \scriptsize
    \setlength{\tabcolsep}{1mm}
    \newcommand{\sz}{0.15}
    \newcommand{\szb}{0.09}
    \newcommand{\szc}{0.11}
    \newcommand{\szd}{0.14}
    \newcommand{\sze}{0.09}
    \newcommand{\rowsep}{7mm}
    \begin{tabular}{cccccc}
        $32^3$, $6 \!\!\times\!\! 6$ & $64^3$, $4 \!\!\times\!\! 4$ & $64^3$, $6 \!\!\times\!\! 6$ & $128^3$, $3 \!\!\times\!\! 3$ & $128^3$, $6 \!\!\times\!\! 6$ & GT \\[5pt]
        \includegraphics[width=\sz\textwidth,trim={0 0 0 50},clip]{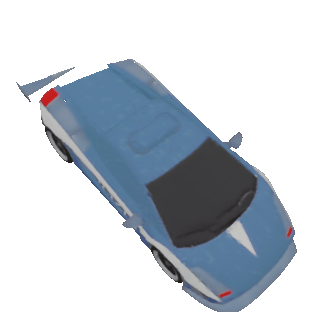} &
        \includegraphics[width=\sz\textwidth,trim={0 0 0 50},clip]{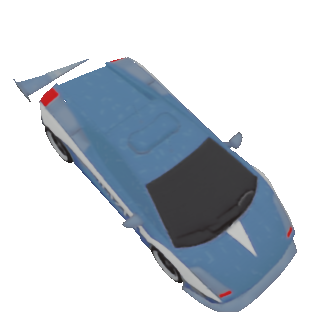} &
        \includegraphics[width=\sz\textwidth,trim={0 0 0 50},clip]{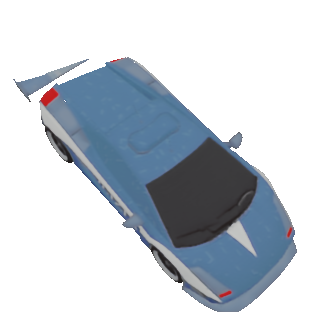} &
        \includegraphics[width=\sz\textwidth,trim={0 0 0 50},clip]{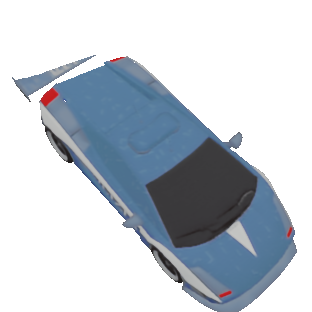} &
        \includegraphics[width=\sz\textwidth,trim={0 0 0 50},clip]{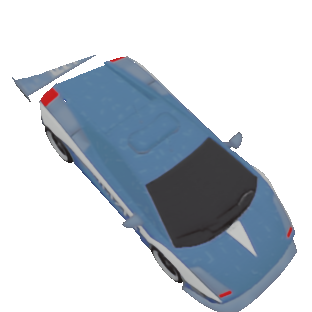} &
        \includegraphics[width=\sz\textwidth,trim={0 0 0 50},clip]{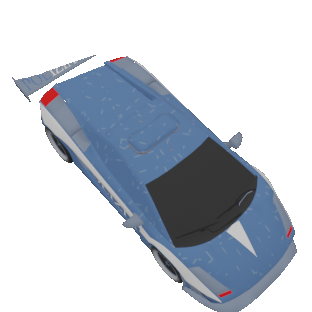} \\[\rowsep]

        \includegraphics[width=\szb\textwidth,trim={90 0 60 85},clip]{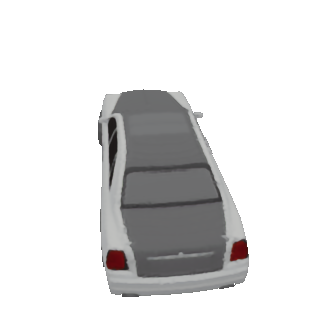} &
        \includegraphics[width=\szb\textwidth,trim={90 0 60 85},clip]{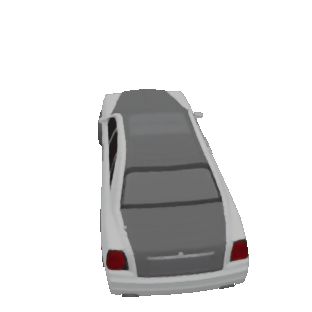} &
        \includegraphics[width=\szb\textwidth,trim={90 0 60 85},clip]{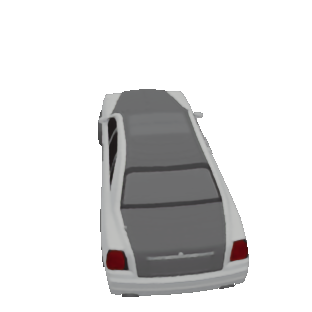} &
        \includegraphics[width=\szb\textwidth,trim={90 0 60 85},clip]{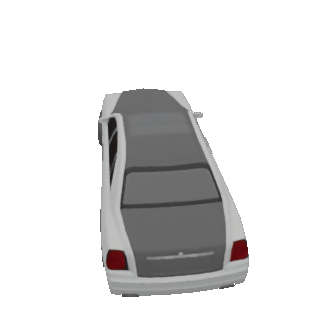} &
        \includegraphics[width=\szb\textwidth,trim={90 0 60 85},clip]{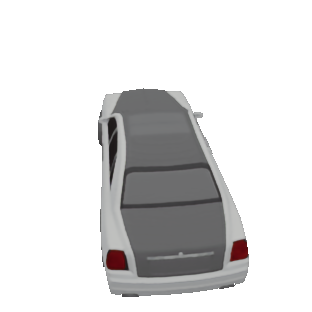} &
        \includegraphics[width=\szb\textwidth,trim={90 0 60 85},clip]{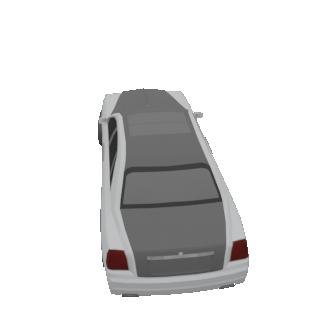} \\[\rowsep]

        \includegraphics[width=\szc\textwidth,trim={70 80 80 120},clip]{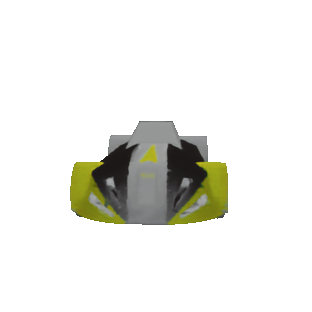} &
        \includegraphics[width=\szc\textwidth,trim={70 80 80 120},clip]{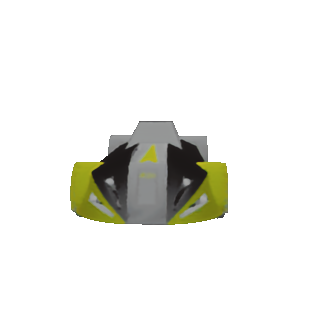} &
        \includegraphics[width=\szc\textwidth,trim={70 80 80 120},clip]{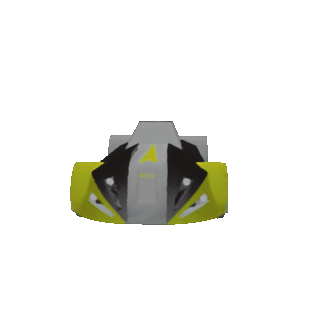} &
        \includegraphics[width=\szc\textwidth,trim={70 80 80 120},clip]{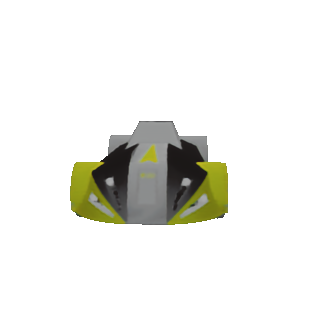} &
        \includegraphics[width=\szc\textwidth,trim={70 80 80 120},clip]{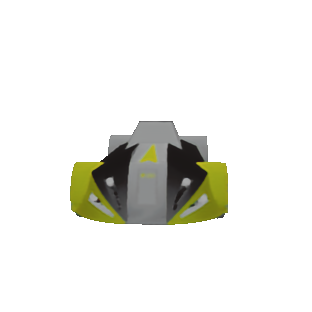} &
        \includegraphics[width=\szc\textwidth,trim={70 80 80 120},clip]{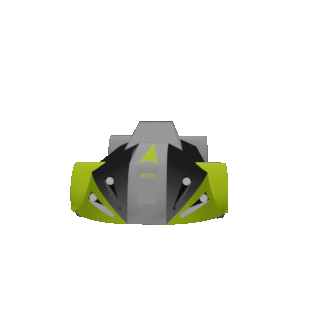} \\[\rowsep]

        \includegraphics[width=\szd\textwidth,trim={0 80 0 95},clip]{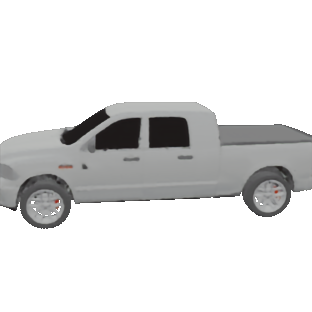} &
        \includegraphics[width=\szd\textwidth,trim={0 80 0 95},clip]{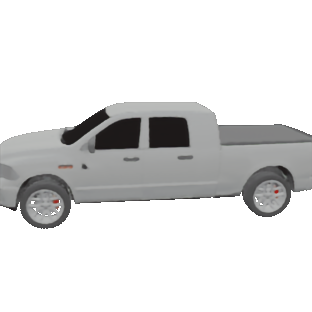} &
        \includegraphics[width=\szd\textwidth,trim={0 80 0 95},clip]{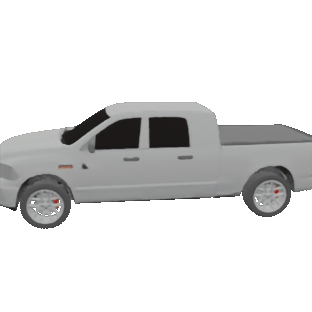} &
        \includegraphics[width=\szd\textwidth,trim={0 80 0 95},clip]{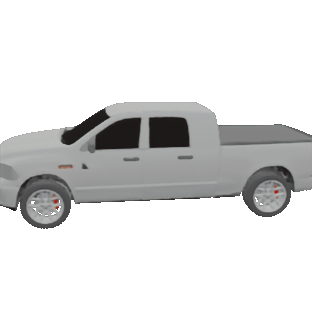} &
        \includegraphics[width=\szd\textwidth,trim={0 80 0 95},clip]{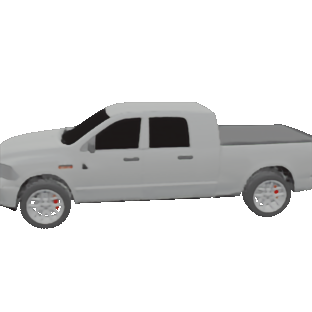} &
        \includegraphics[width=\szd\textwidth,trim={0 80 0 95},clip]{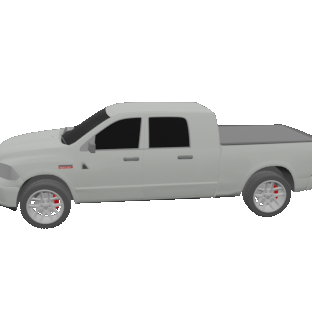} \\[\rowsep]

        \includegraphics[width=\sze\textwidth,trim={65 0 80 70},clip]{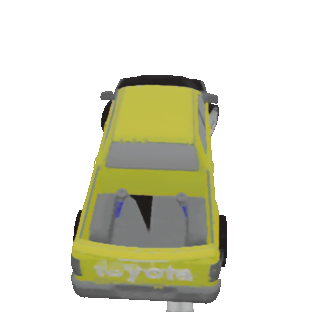} &
        \includegraphics[width=\sze\textwidth,trim={65 0 80 70},clip]{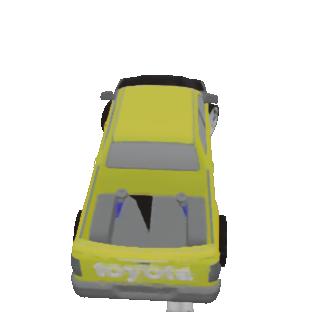} &
        \includegraphics[width=\sze\textwidth,trim={65 0 80 70},clip]{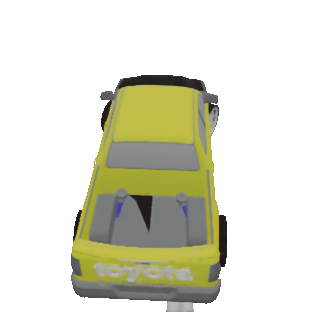} &
        \includegraphics[width=\sze\textwidth,trim={65 0 80 70},clip]{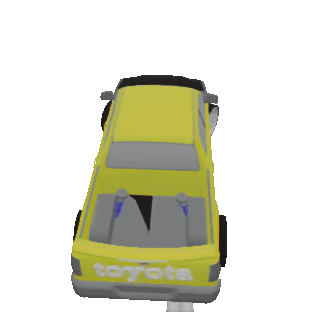} &
        \includegraphics[width=\sze\textwidth,trim={65 0 80 70},clip]{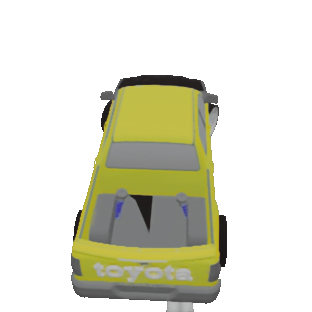} &
        \includegraphics[width=\sze\textwidth,trim={65 0 80 70},clip]{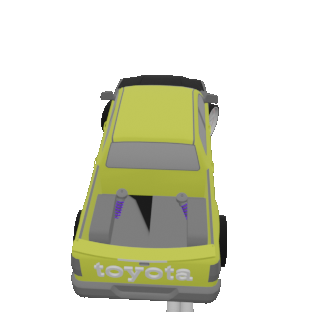} \\[\rowsep]

        \includegraphics[width=\szd\textwidth,trim={0 55 0 90},clip]{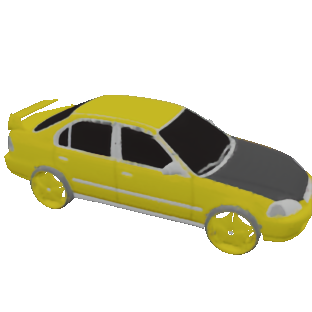} &
        \includegraphics[width=\szd\textwidth,trim={0 55 0 90},clip]{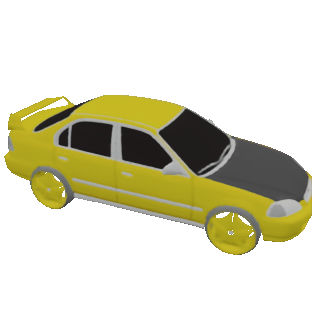} &
        \includegraphics[width=\szd\textwidth,trim={0 55 0 90},clip]{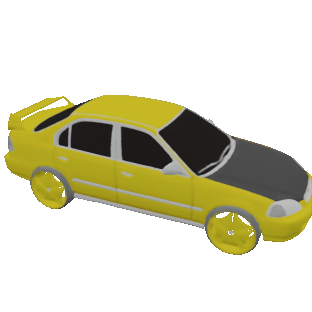} &
        \includegraphics[width=\szd\textwidth,trim={0 55 0 90},clip]{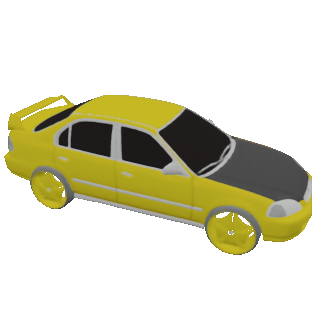} &
        \includegraphics[width=\szd\textwidth,trim={0 55 0 90},clip]{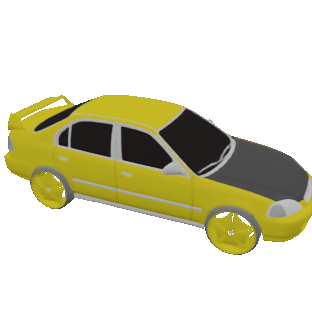} &
        \includegraphics[width=\szd\textwidth,trim={0 55 0 90},clip]{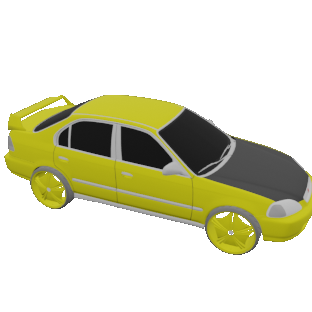} \\[\rowsep]

        \includegraphics[width=\szd\textwidth,trim={0 75 0 100},clip]{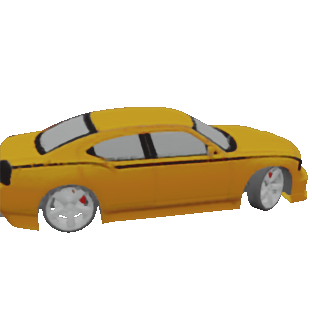} &
        \includegraphics[width=\szd\textwidth,trim={0 75 0 100},clip]{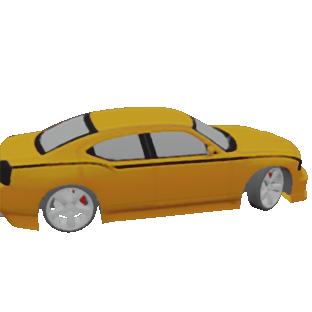} &
        \includegraphics[width=\szd\textwidth,trim={0 75 0 100},clip]{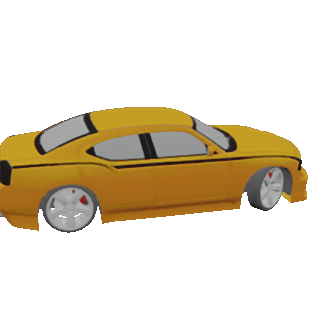} &
        \includegraphics[width=\szd\textwidth,trim={0 75 0 100},clip]{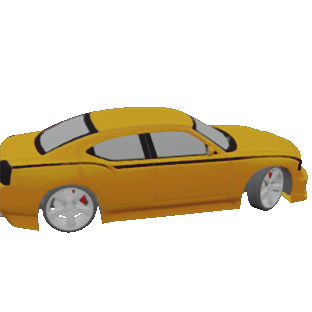} &
        \includegraphics[width=\szd\textwidth,trim={0 75 0 100},clip]{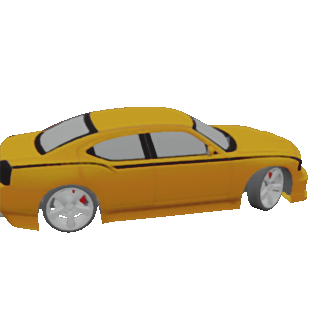} &
        \includegraphics[width=\szd\textwidth,trim={0 75 0 100},clip]{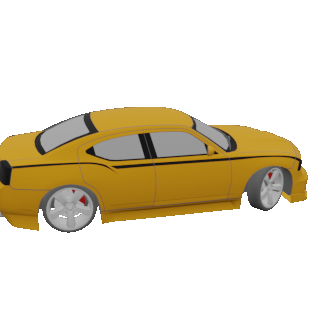} \\

    \end{tabular}
    \caption{\textbf{Qualitative results of our model on unseen ShapeNet~\cite{chang2015shapenet} car scenes for different DeepSurfel parameters.}
    The column names denote DeepSurfel grid and patch resolution respectively. We used DeepSurfels with 3 feature and 3 color channels (3+3 configuration).
    A quantitative comparison is given in \tablename~\ref{tab:app:ablation_study}.
    }
    \label{fig:app:nvs_ch3}
\end{figure*}

\begin{figure*}
    \centering
    \scriptsize
    \setlength{\tabcolsep}{1mm}
    \newcommand{\sz}{0.15}
    \newcommand{\szb}{0.09}
    \newcommand{\szc}{0.11}
    \newcommand{\szd}{0.14}
    \newcommand{\sze}{0.09}
    \newcommand{\rowsep}{7mm}
    \begin{tabular}{cccccc}
        $32^3$, $6 \!\!\times\!\! 6$ & $64^3$, $4 \!\!\times\!\! 4$ & $64^3$, $6 \!\!\times\!\! 6$ & $128^3$, $3 \!\!\times\!\! 3$ & $128^3$, $6 \!\!\times\!\! 6$ & GT \\[5pt]
        \includegraphics[width=\sz\textwidth,trim={0 0 0 50},clip]{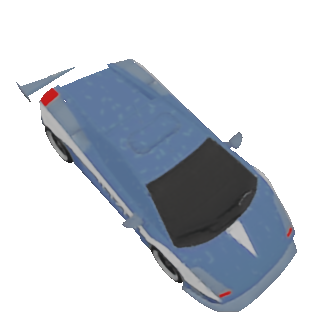} &
        \includegraphics[width=\sz\textwidth,trim={0 0 0 50},clip]{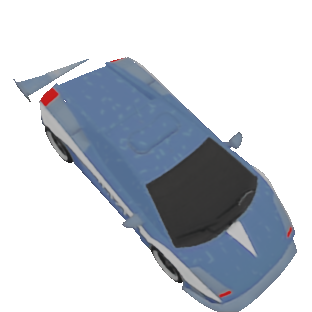} &
        \includegraphics[width=\sz\textwidth,trim={0 0 0 50},clip]{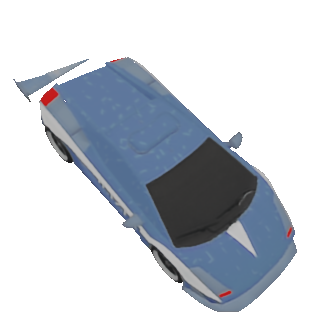} &
        \includegraphics[width=\sz\textwidth,trim={0 0 0 50},clip]{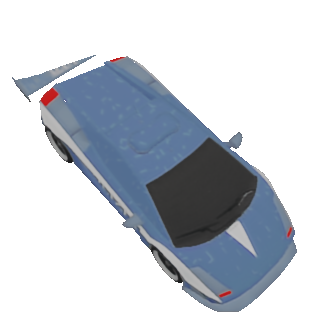} &
        \includegraphics[width=\sz\textwidth,trim={0 0 0 50},clip]{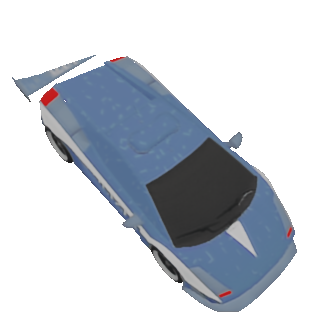} &
        \includegraphics[width=\sz\textwidth,trim={0 0 0 50},clip]{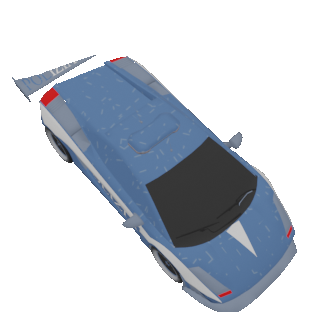} \\[\rowsep]

        \includegraphics[width=\szb\textwidth,trim={90 0 60 85},clip]{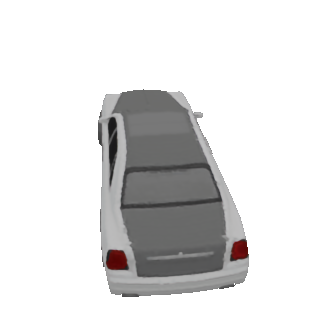} &
        \includegraphics[width=\szb\textwidth,trim={90 0 60 85},clip]{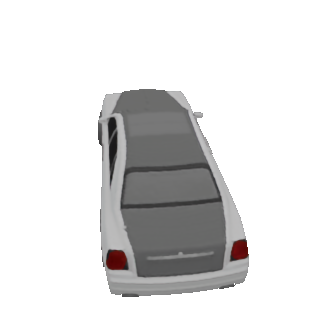} &
        \includegraphics[width=\szb\textwidth,trim={90 0 60 85},clip]{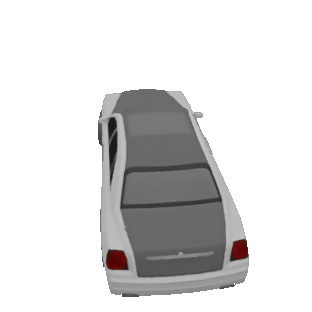} &
        \includegraphics[width=\szb\textwidth,trim={90 0 60 85},clip]{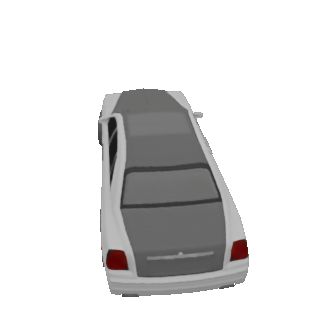} &
        \includegraphics[width=\szb\textwidth,trim={90 0 60 85},clip]{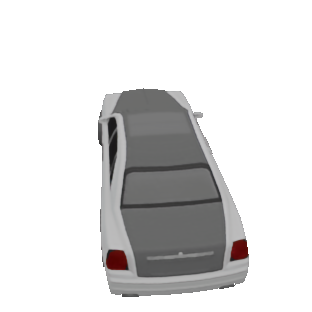} &
        \includegraphics[width=\szb\textwidth,trim={90 0 60 85},clip]{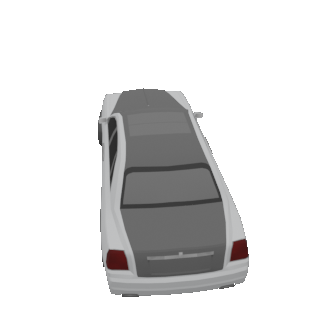} \\[\rowsep]

        \includegraphics[width=\szc\textwidth,trim={70 80 80 120},clip]{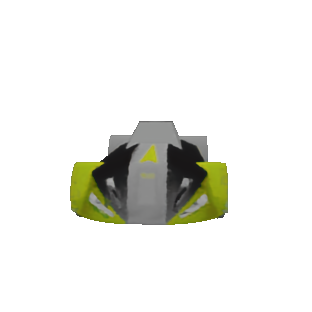} &
        \includegraphics[width=\szc\textwidth,trim={70 80 80 120},clip]{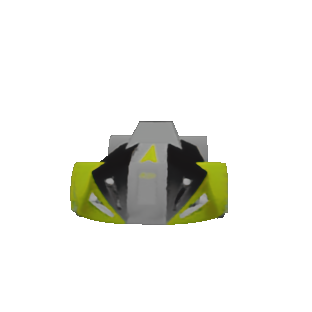} &
        \includegraphics[width=\szc\textwidth,trim={70 80 80 120},clip]{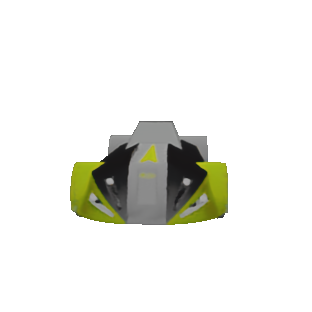} &
        \includegraphics[width=\szc\textwidth,trim={70 80 80 120},clip]{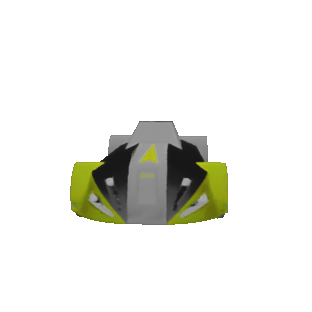} &
        \includegraphics[width=\szc\textwidth,trim={70 80 80 120},clip]{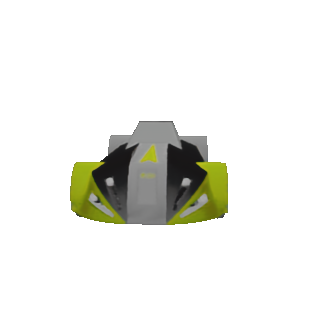} &
        \includegraphics[width=\szc\textwidth,trim={70 80 80 120},clip]{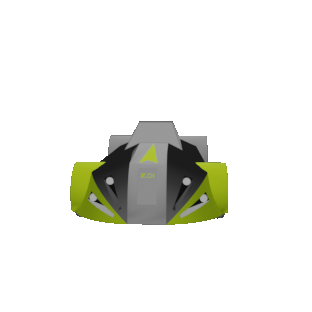} \\[\rowsep]

        \includegraphics[width=\szd\textwidth,trim={0 80 0 95},clip]{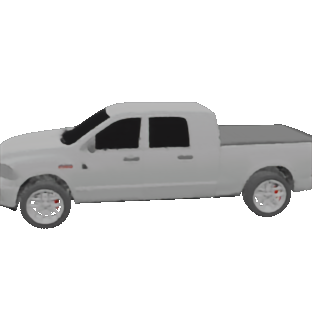} &
        \includegraphics[width=\szd\textwidth,trim={0 80 0 95},clip]{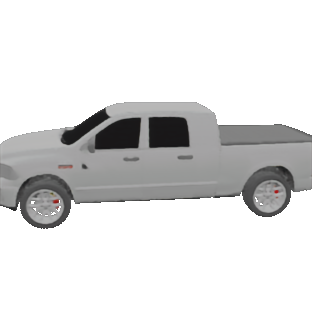} &
        \includegraphics[width=\szd\textwidth,trim={0 80 0 95},clip]{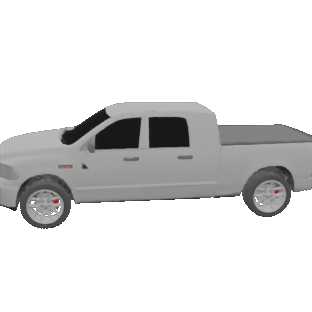} &
        \includegraphics[width=\szd\textwidth,trim={0 80 0 95},clip]{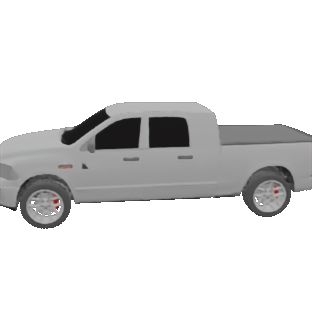} &
        \includegraphics[width=\szd\textwidth,trim={0 80 0 95},clip]{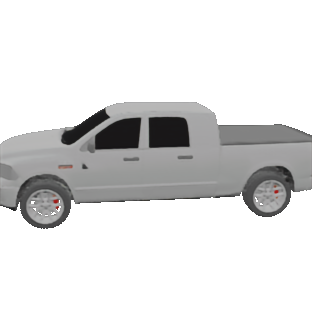} &
        \includegraphics[width=\szd\textwidth,trim={0 80 0 95},clip]{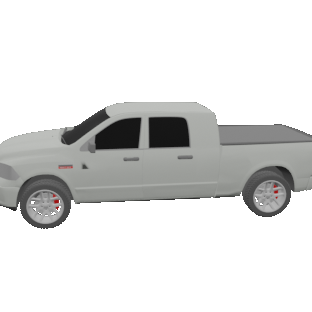} \\[\rowsep]

        \includegraphics[width=\sze\textwidth,trim={65 0 80 70},clip]{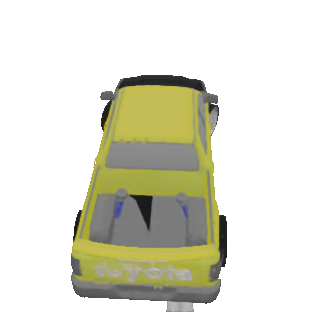} &
        \includegraphics[width=\sze\textwidth,trim={65 0 80 70},clip]{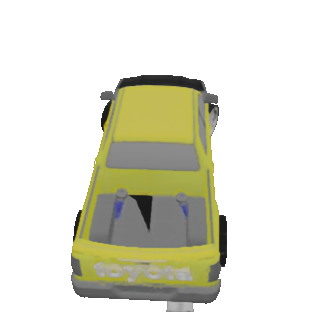} &
        \includegraphics[width=\sze\textwidth,trim={65 0 80 70},clip]{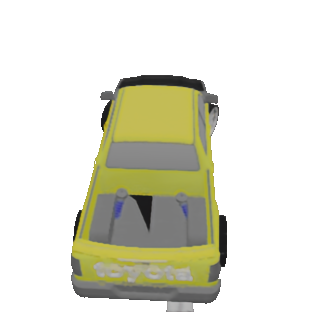} &
        \includegraphics[width=\sze\textwidth,trim={65 0 80 70},clip]{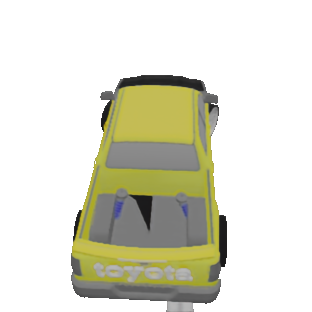} &
        \includegraphics[width=\sze\textwidth,trim={65 0 80 70},clip]{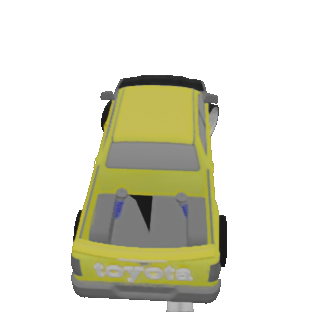} &
        \includegraphics[width=\sze\textwidth,trim={65 0 80 70},clip]{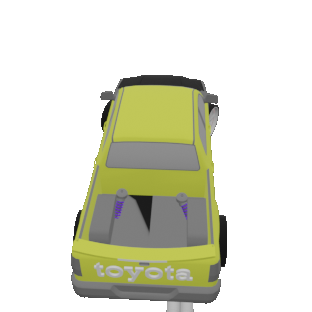} \\[\rowsep]

        \includegraphics[width=\szd\textwidth,trim={0 55 0 90},clip]{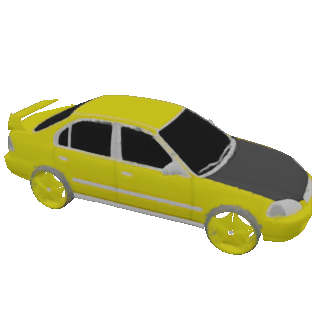} &
        \includegraphics[width=\szd\textwidth,trim={0 55 0 90},clip]{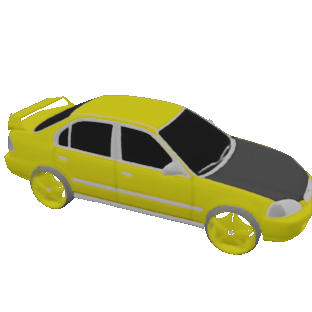} &
        \includegraphics[width=\szd\textwidth,trim={0 55 0 90},clip]{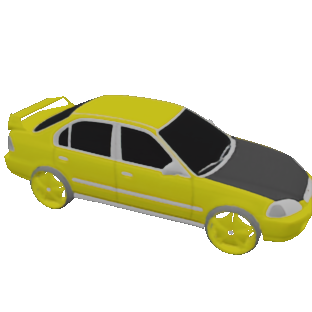} &
        \includegraphics[width=\szd\textwidth,trim={0 55 0 90},clip]{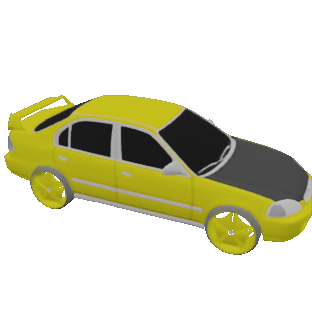} &
        \includegraphics[width=\szd\textwidth,trim={0 55 0 90},clip]{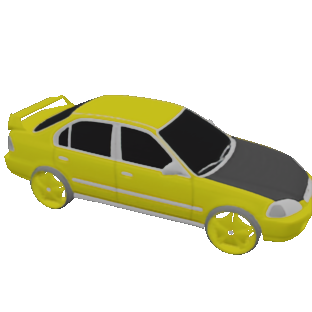} &
        \includegraphics[width=\szd\textwidth,trim={0 55 0 90},clip]{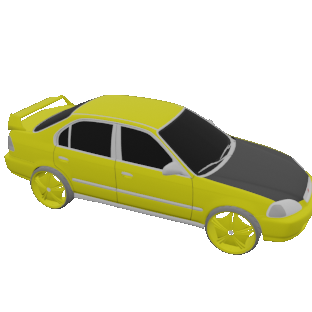} \\[\rowsep]

        \includegraphics[width=\szd\textwidth,trim={0 75 0 100},clip]{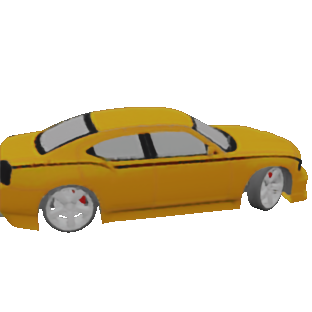} &
        \includegraphics[width=\szd\textwidth,trim={0 75 0 100},clip]{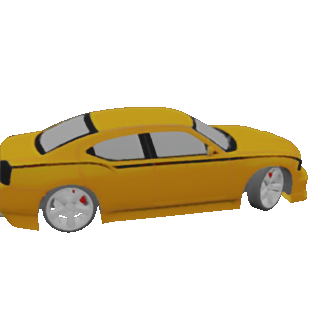} &
        \includegraphics[width=\szd\textwidth,trim={0 75 0 100},clip]{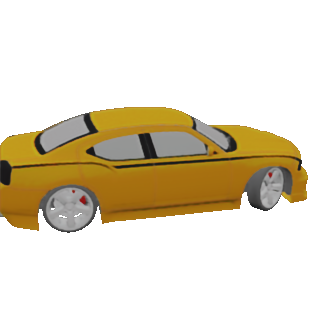} &
        \includegraphics[width=\szd\textwidth,trim={0 75 0 100},clip]{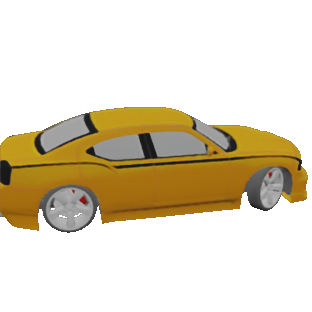} &
        \includegraphics[width=\szd\textwidth,trim={0 75 0 100},clip]{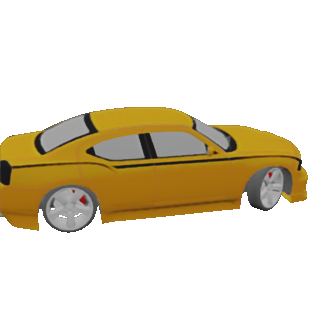} &
        \includegraphics[width=\szd\textwidth,trim={0 75 0 100},clip]{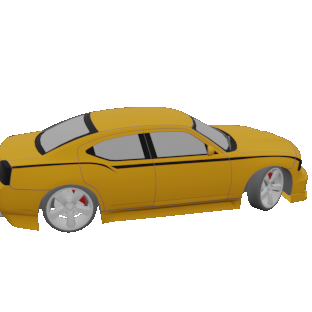} \\

    \end{tabular}
    \caption{\textbf{Qualitative results of our model on unseen ShapeNet~\cite{chang2015shapenet} car scenes for different DeepSurfel parameters.}
    DeepSurfels with 5 feature and 3 color channels (5+3 configuration) demonstrate better results compared to our method with less channels (3+3) displayed in \figurename~\ref{fig:app:nvs_ch3}.
    Quantitative comparison is given in \tablename~\ref{tab:app:ablation_study}.
    The column name denotes DeepSurfel gird and patch resolution respectively.
    }
    \label{fig:app:nvs_ch5}
\end{figure*}

Quantitative and qualitative results (\tablename~\ref{tab:app:ablation_study}, \figurename~\ref{fig:app:nvs_ch3} and \ref{fig:app:nvs_ch5}) demonstrate that additional two feature channels are beneficial for the quality of rendered images.


%% file: ms.bbl
\begin{thebibliography}{100}\itemsep=-1pt

\bibitem{achlioptas2017learning}
Panos Achlioptas, Olga Diamanti, Ioannis Mitliagkas, and Leonidas Guibas.
\newblock Learning representations and generative models for 3d point clouds.
\newblock In {\em Proc.~International Conference on Machine Learning (ICML)},
  2018.

\bibitem{aliev2019neural}
Kara-Ali Aliev, Artem Sevastopolsky, Maria Kolos, Dmitry Ulyanov, and Victor
  Lempitsky.
\newblock Neural point-based graphics.
\newblock In {\em Proc.~European Conference on Computer Vision (ECCV)}, 2020.

\bibitem{Allene-et-al-ICPR-2008}
C{\'e}dric All{\`e}ne, Jean-Philippe Pons, and Renaud Keriven.
\newblock Seamless image-based texture atlases using multi-band blending.
\newblock In {\em Proc.~International Conference on Computer Vision (ICCV)},
  2008.

\bibitem{Armando-et-al-3DV-2019}
Matthieu Armando, Jean{-}S{\'{e}}bastien Franco, and Edmond Boyer.
\newblock Adaptive mesh texture for multi-view appearance modeling.
\newblock In {\em Proc.~International Conference on 3D Vision (3DV)}, 2019.

\bibitem{Bernardini-et-al-TVCG-2001}
Fausto Bernardini, Ioana~M. Martin, and Holly~E. Rushmeier.
\newblock High-quality texture reconstruction from multiple scans.
\newblock {\em IEEE Transactions on Visualization and Computer Graphics}, 2001.

\bibitem{Bi-et-al-TOG-2017}
Sai Bi, Nima~Khademi Kalantari, and Ravi Ramamoorthi.
\newblock Patch-based optimization for image-based texture mapping.
\newblock {\em ACM Transactions on Graphics}, 2017.

\bibitem{Bi-et-al-ECCV-2020}
Sai Bi, Zexiang Xu, Kalyan Sunkavalli, Miloš Hašan, Yannick Hold-Geoffroy,
  David Kriegman, and Ravi Ramamoorthi.
\newblock Deep reflectance volumes: Relightable reconstructions from multi-view
  photometric images.
\newblock In {\em Proc.~European Conference on Computer Vision (ECCV)}, 2020.

\bibitem{bircher2015structural}
Andreas Bircher, Kostas Alexis, Michael Burri, Philipp Oettershagen, Sammy
  Omari, Thomas Mantel, and Roland Siegwart.
\newblock Structural inspection path planning via iterative viewpoint
  resampling with application to aerial robotics.
\newblock In {\em IEEE International Conference on Robotics and Automation},
  2015.

\bibitem{breitenmoser2012surface}
Andreas Breitenmoser and Roland Siegwart.
\newblock Surface reconstruction and path planning for industrial inspection
  with a climbing robot.
\newblock In {\em Proc.~International Conference on Applied Robotics for the
  Power Industry (CARPI)}, 2012.

\bibitem{brock2016generative}
Andrew Brock, Theodore Lim, James~M Ritchie, and Nick Weston.
\newblock Generative and discriminative voxel modeling with convolutional
  neural networks.
\newblock {\em arXiv preprint arXiv:1608.04236}, 2016.

\bibitem{chabra2020deep}
Rohan Chabra, Jan~Eric Lenssen, Eddy Ilg, Tanner Schmidt, Julian Straub, Steven
  Lovegrove, and Richard Newcombe.
\newblock Deep local shapes: Learning local sdf priors for detailed 3d
  reconstruction.
\newblock In {\em Proc.~European Conference on Computer Vision (ECCV)}, 2020.

\bibitem{chang2015shapenet}
Angel~X Chang, Thomas Funkhouser, Leonidas Guibas, Pat Hanrahan, Qixing Huang,
  Zimo Li, Silvio Savarese, Manolis Savva, Shuran Song, Hao Su, et~al.
\newblock Shapenet: An information-rich 3d model repository.
\newblock {\em arXiv preprint arXiv:1512.03012}, 2015.

\bibitem{Chen-et-al-CVPR-2019}
Zhiqin Chen and Hao Zhang.
\newblock Learning implicit fields for generative shape modeling.
\newblock In {\em Proc.~International Conference on Computer Vision and Pattern
  Recognition (CVPR)}, 2019.

\bibitem{Chu-et-al-ECCV-2020}
Hang Chu, Shugao Ma, Fernando~De la Torre, Sanja Fidler, and Yaser Sheikh.
\newblock Expressive telepresence via modular codec avatars.
\newblock In {\em Proc.~European Conference on Computer Vision (ECCV)}, Lecture
  Notes in Computer Science, 2020.

\bibitem{blender20}
Blender~Online Community.
\newblock {\em Blender - a 3D modelling and rendering package}.
\newblock Blender Foundation, Stichting Blender Foundation, Amsterdam, 2020.

\bibitem{Cook-et-al-SIGGRAPH-1984}
Robert~L. Cook, Thomas Porter, and Loren Carpenter.
\newblock Distributed ray tracing.
\newblock In {\em ACM Transactions on Graphics (Proc.~SIGGRAPH)}, 1984.

\bibitem{curless1996volumetric}
Brian Curless and Marc Levoy.
\newblock A volumetric method for building complex models from range images.
\newblock In {\em ACM Transactions on Graphics (Proc.~SIGGRAPH)}, 1996.

\bibitem{debevec1996modeling}
Paul~E Debevec, Camillo~J Taylor, and Jitendra Malik.
\newblock Modeling and rendering architecture from photographs: A hybrid
  geometry-and image-based approach.
\newblock In {\em ACM Transactions on Graphics (Proc.~SIGGRAPH)}, 1996.

\bibitem{Eisemann-et-al-CGF-2008}
M. Eisemann, B. {De Decker}, M. Magnor, P. Bekaert, E. de Aguiar, N. Ahmed, C.
  Theobalt, and A. Sellent.
\newblock {Floating Textures}.
\newblock {\em Computer Graphics Forum}, 2008.

\bibitem{fan2017point}
Haoqiang Fan, Hao Su, and Leonidas~J Guibas.
\newblock A point set generation network for 3d object reconstruction from a
  single image.
\newblock In {\em Proc.~International Conference on Computer Vision and Pattern
  Recognition (CVPR)}, 2017.

\bibitem{flynn2019deepview}
John Flynn, Michael Broxton, Paul Debevec, Matthew DuVall, Graham Fyffe, Ryan
  Overbeck, Noah Snavely, and Richard Tucker.
\newblock Deepview: View synthesis with learned gradient descent.
\newblock In {\em Proc.~International Conference on Computer Vision and Pattern
  Recognition (CVPR)}, 2019.

\bibitem{Fu-et-al-CVPR-2020}
Yanping Fu, Qingan Yan, Jie Liao, and Chunxia Xiao.
\newblock Joint texture and geometry optimization for {RGB-D} reconstruction.
\newblock In {\em Proc.~International Conference on Computer Vision and Pattern
  Recognition (CVPR)}, 2020.

\bibitem{Fu-et-al-CVPR-2018}
Yanping Fu, Qingan Yan, Long Yang, Jie Liao, and Chunxia Xiao.
\newblock Texture mapping for 3d reconstruction with {RGB-D} sensor.
\newblock In {\em Proc.~International Conference on Computer Vision and Pattern
  Recognition (CVPR)}, 2018.

\bibitem{gadelha20173d}
Matheus Gadelha, Subhransu Maji, and Rui Wang.
\newblock 3d shape induction from 2d views of multiple objects.
\newblock In {\em Proc.~International Conference on 3D Vision (3DV)}, 2017.

\bibitem{Gal-et-al-CGF-2010}
Ran Gal, Yonatan Wexler, Eyal Ofek, Hugues Hoppe, and Daniel Cohen{-}Or.
\newblock Seamless montage for texturing models.
\newblock {\em Computer Graphics Forum}, 2010.

\bibitem{garrido2013application}
Santiago Garrido, Mar{\'\i}a Malfaz, and Dolores Blanco.
\newblock Application of the fast marching method for outdoor motion planning
  in robotics.
\newblock {\em Robotics and Autonomous Systems}, 2013.

\bibitem{Goldluecke-et-al-IJCV-2014}
Bastian Goldl{\"{u}}cke, Mathieu Aubry, Kalin Kolev, and Daniel Cremers.
\newblock A super-resolution framework for high-accuracy multiview
  reconstruction.
\newblock {\em International Journal of Computer Vision}, 2014.

\bibitem{hane20173d}
Christian H{\"a}ne, Lionel Heng, Gim~Hee Lee, Friedrich Fraundorfer, Paul
  Furgale, Torsten Sattler, and Marc Pollefeys.
\newblock 3d visual perception for self-driving cars using a multi-camera
  system: Calibration, mapping, localization, and obstacle detection.
\newblock {\em Image and Vision Computing}, 2017.

\bibitem{Hanocka2020p2m}
Rana Hanocka, Gal Metzer, Raja Giryes, and Daniel Cohen-Or.
\newblock Point2mesh: A self-prior for deformable meshes.
\newblock {\em ACM Transactions on Graphics}, 2020.

\bibitem{huang2020advtex}
Jingwei Huang, Justus Thies, Angela Dai, Abhijit Kundu, Chiyu Jiang, Leonidas~J
  Guibas, Matthias Nie{\ss}ner, and Thomas Funkhouser.
\newblock Adversarial texture optimization from {RGB-D} scans.
\newblock In {\em Proc.~International Conference on Computer Vision and Pattern
  Recognition (CVPR)}, 2020.

\bibitem{Chiyu-et-al-CVPR-2020}
Chiyu Jiang, Avneesh Sud, Ameesh Makadia, Jingwei Huang, Matthias Nie{\ss}ner,
  and Thomas Funkhouser.
\newblock Local implicit grid representations for 3d scenes.
\newblock In {\em Proc.~International Conference on Computer Vision and Pattern
  Recognition (CVPR)}, 2020.

\bibitem{kahler2015very}
Olaf K{\"a}hler, Victor~Adrian Prisacariu, Carl~Yuheng Ren, Xin Sun, Philip
  Torr, and David Murray.
\newblock Very high frame rate volumetric integration of depth images on mobile
  devices.
\newblock {\em IEEE Transactions on Visualization and Computer Graphics}, 2015.

\bibitem{kanazawa2018learning}
Angjoo Kanazawa, Shubham Tulsiani, Alexei~A Efros, and Jitendra Malik.
\newblock Learning category-specific mesh reconstruction from image
  collections.
\newblock In {\em Proc.~European Conference on Computer Vision (ECCV)}, 2018.

\bibitem{kingma2014adam}
Diederik~P. Kingma and Jimmy Ba.
\newblock Adam: {A} method for stochastic optimization.
\newblock In {\em Proc.~International Conference on Learning Representations
  (ICLR)}, 2015.

\bibitem{kutulakos1999theory}
Kiriakos~N Kutulakos and Steven~M Seitz.
\newblock A theory of shape by space carving.
\newblock In {\em Proc.~International Conference on Computer Vision (ICCV)},
  1999.

\bibitem{lee2020texturefusion}
Joo~Ho Lee, Hyunho Ha, Yue Dong, Xin Tong, and Min~H Kim.
\newblock Texturefusion: High-quality texture acquisition for real-time rgb-d
  scanning.
\newblock In {\em Proc.~International Conference on Computer Vision and Pattern
  Recognition (CVPR)}, 2020.

\bibitem{Lempitsky-Ivanov-CVPR-2007}
Victor Lempitsky and Denis Ivanov.
\newblock Seamless mosaicing of image-based texture maps.
\newblock In {\em Proc.~International Conference on Computer Vision and Pattern
  Recognition (CVPR)}, 2007.

\bibitem{Lensch-et-al-GM-2001}
Hendrik P.~A. Lensch, Wolfgang Heidrich, and Hans-Peter Seidel.
\newblock A silhouette-based algorithm for texture registration and stitching.
\newblock {\em Graphical Models}, 2001.

\bibitem{Li-et-al-CVPR-2019}
Yawei Li, Vagia Tsiminaki, Radu Timofte, Marc Pollefeys, and Luc~Van Gool.
\newblock 3d appearance super-resolution with deep learning.
\newblock In {\em Proc.~International Conference on Computer Vision and Pattern
  Recognition (CVPR)}, 2019.

\bibitem{liao2018deep}
Yiyi Liao, Simon Donne, and Andreas Geiger.
\newblock Deep marching cubes: Learning explicit surface representations.
\newblock In {\em Proc.~International Conference on Computer Vision and Pattern
  Recognition (CVPR)}, 2018.

\bibitem{liu2019soft}
Shichen Liu, Tianye Li, Weikai Chen, and Hao Li.
\newblock Soft rasterizer: A differentiable renderer for image-based 3d
  reasoning.
\newblock In {\em Proc.~International Conference on Computer Vision (ICCV)},
  2019.

\bibitem{Liu-et-al-TVCG-2019}
Z. {Liu}, Y. {Cao}, Z. {Kuang}, L. {Kobbelt}, and S. {Hu}.
\newblock High-quality textured 3d shape reconstruction with cascaded fully
  convolutional networks.
\newblock {\em IEEE Transactions on Visualization and Computer Graphics}, 2019.

\bibitem{Lombard-et-al-TOG-2018}
Stephen Lombardi, Jason~M. Saragih, Tomas Simon, and Yaser Sheikh.
\newblock Deep appearance models for face rendering.
\newblock {\em ACM Transactions on Graphics}, 2018.

\bibitem{lombardi2019neural}
Stephen Lombardi, Tomas Simon, Jason Saragih, Gabriel Schwartz, Andreas
  Lehrmann, and Yaser Sheikh.
\newblock Neural volumes: Learning dynamic renderable volumes from images.
\newblock {\em ACM Transactions on Graphics}, 2019.

\bibitem{Maier-et-al-ICCV-2017}
Robert Maier, Kihwan Kim, Daniel Cremers, Jan Kautz, and Matthias Nie{\ss}ner.
\newblock Intrinsic3d: High-quality 3d reconstruction by joint appearance and
  geometry optimization with spatially-varying lighting.
\newblock In {\em Proc.~International Conference on Computer Vision (ICCV)},
  2017.

\bibitem{Mescheder-et-al-CVPR-2019}
Lars~M. Mescheder, Michael Oechsle, Michael Niemeyer, Sebastian Nowozin, and
  Andreas Geiger.
\newblock Occupancy networks: Learning 3d reconstruction in function space.
\newblock In {\em Proc.~International Conference on Computer Vision and Pattern
  Recognition (CVPR)}, 2019.

\bibitem{LEAP:CVPR:21}
Marko Mihajlovic, Yan Zhang, Michael~J. Black, and Siyu Tang.
\newblock {LEAP}: Learning articulated occupancy of people.
\newblock In {\em Proc.~International Conference on Computer Vision and Pattern
  Recognition (CVPR)}, 2021.

\bibitem{mildenhall2019local}
Ben Mildenhall, Pratul~P Srinivasan, Rodrigo Ortiz-Cayon, Nima~Khademi
  Kalantari, Ravi Ramamoorthi, Ren Ng, and Abhishek Kar.
\newblock Local light field fusion: Practical view synthesis with prescriptive
  sampling guidelines.
\newblock {\em ACM Transactions on Graphics}, 2019.

\bibitem{mildenhall2020nerf}
Ben Mildenhall, Pratul~P Srinivasan, Matthew Tancik, Jonathan~T Barron, Ravi
  Ramamoorthi, and Ren Ng.
\newblock {NeRF}: Representing scenes as neural radiance fields for view
  synthesis.
\newblock In {\em Proc.~European Conference on Computer Vision (ECCV)}, 2020.

\bibitem{KinectFusion}
Richard~A Newcombe, Shahram Izadi, Otmar Hilliges, David Molyneaux, David Kim,
  Andrew~J Davison, Pushmeet Kohi, Jamie Shotton, Steve Hodges, and Andrew
  Fitzgibbon.
\newblock {KinectFusion}: Real-time dense surface mapping and tracking.
\newblock In {\em IEEE International Symposium on Mixed and Augmented Reality},
  2011.

\bibitem{newcombe2011dtam}
Richard~A Newcombe, Steven~J Lovegrove, and Andrew~J Davison.
\newblock Dtam: Dense tracking and mapping in real-time.
\newblock In {\em Proc.~International Conference on Computer Vision (ICCV)},
  2011.

\bibitem{DVR}
Michael Niemeyer, Lars Mescheder, Michael Oechsle, and Andreas Geiger.
\newblock Differentiable volumetric rendering: Learning implicit 3d
  representations without 3d supervision.
\newblock In {\em Proc.~International Conference on Computer Vision and Pattern
  Recognition (CVPR)}, 2020.

\bibitem{niessner2013real}
Matthias Nie{\ss}ner, Michael Zollh{\"o}fer, Shahram Izadi, and Marc
  Stamminger.
\newblock Real-time 3d reconstruction at scale using voxel hashing.
\newblock {\em ACM Transactions on Graphics}, 2013.

\bibitem{Oechsle-et-al-ICCV-2019}
Michael Oechsle, Lars Mescheder, Michael Niemeyer, Thilo Strauss, and Andreas
  Geiger.
\newblock Texture fields: Learning texture representations in function space.
\newblock In {\em Proc.~International Conference on Computer Vision (ICCV)},
  2019.

\bibitem{oechsle2020learning}
Michael Oechsle, Michael Niemeyer, Christian Reiser, Lars Mescheder, Thilo
  Strauss, and Andreas Geiger.
\newblock Learning implicit surface light fields.
\newblock In {\em Proc.~International Conference on 3D Vision (3DV)}, 2020.

\bibitem{Park-et-al-CVPR-2019}
Jeong~Joon Park, Peter Florence, Julian Straub, Richard~A. Newcombe, and Steven
  Lovegrove.
\newblock Deepsdf: Learning continuous signed distance functions for shape
  representation.
\newblock In {\em Proc.~International Conference on Computer Vision and Pattern
  Recognition (CVPR)}, 2019.

\bibitem{peng2020convolutional}
Songyou Peng, Michael Niemeyer, Lars Mescheder, Marc Pollefeys, and Andreas
  Geiger.
\newblock Convolutional occupancy networks.
\newblock In {\em Proc.~European Conference on Computer Vision (ECCV)}, 2020.

\bibitem{penner2017soft}
Eric Penner and Li Zhang.
\newblock Soft 3d reconstruction for view synthesis.
\newblock {\em ACM Transactions on Graphics}, 2017.

\bibitem{pfister2000surfels}
Hanspeter Pfister, Matthias Zwicker, Jeroen Van~Baar, and Markus Gross.
\newblock {Surfels}: Surface elements as rendering primitives.
\newblock In {\em ACM Transactions on Graphics (Proc.~SIGGRAPH)}, 2000.

\bibitem{rahaman2018spectral}
Nasim Rahaman, Aristide Baratin, Devansh Arpit, Felix Draxler, Min Lin, Fred
  Hamprecht, Yoshua Bengio, and Aaron Courville.
\newblock On the spectral bias of neural networks.
\newblock In {\em Proc.~International Conference on Machine Learning (ICML)},
  2019.

\bibitem{rematas2020neural}
Konstantinos Rematas and Vittorio Ferrari.
\newblock Neural voxel renderer: Learning an accurate and controllable
  rendering tool.
\newblock In {\em Proc.~International Conference on Computer Vision and Pattern
  Recognition (CVPR)}, 2020.

\bibitem{rezende2016unsupervised}
Danilo~Jimenez Rezende, SM~Ali Eslami, Shakir Mohamed, Peter Battaglia, Max
  Jaderberg, and Nicolas Heess.
\newblock Unsupervised learning of 3d structure from images.
\newblock In {\em Proc.~Neural Information Processing Systems (NeurIPS)}, 2016.

\bibitem{Richard-et-al-3DV2019}
Audrey Richard, Ian Cherabier, Martin~R. Oswald, Vagia Tsiminaki, Marc
  Pollefeys, and Konrad Schindler.
\newblock Learned multi-view texture super-resolution.
\newblock In {\em Proc.~International Conference on 3D Vision (3DV)}, 2019.

\bibitem{Riegler2020FVS}
Gernot Riegler and Vladlen Koltun.
\newblock Free view synthesis.
\newblock In {\em Proc.~European Conference on Computer Vision (ECCV)}, 2020.

\bibitem{saito2019pifu}
Shunsuke Saito, Zeng Huang, Ryota Natsume, Shigeo Morishima, Angjoo Kanazawa,
  and Hao Li.
\newblock {PIFu}: Pixel-aligned implicit function for high-resolution clothed
  human digitization.
\newblock In {\em Proc.~International Conference on Computer Vision (ICCV)},
  2019.

\bibitem{saito2020pifuhd}
Shunsuke Saito, Tomas Simon, Jason Saragih, and Hanbyul Joo.
\newblock {PIFuHD}: Multi-level pixel-aligned implicit function for
  high-resolution 3d human digitization.
\newblock In {\em Proc.~International Conference on Computer Vision and Pattern
  Recognition (CVPR)}, 2020.

\bibitem{habitat19iccv}
Manolis Savva, Abhishek Kadian, Oleksandr Maksymets, Yili Zhao, Erik Wijmans,
  Bhavana Jain, Julian Straub, Jia Liu, Vladlen Koltun, Jitendra Malik, Devi
  Parikh, and Dhruv Batra.
\newblock Habitat: {A} {P}latform for {E}mbodied {AI} {R}esearch.
\newblock In {\em Proc.~International Conference on Computer Vision (ICCV)},
  2019.

\bibitem{schops2017real}
Thomas Sch{\"o}ps, Martin~R Oswald, Pablo Speciale, Shuoran Yang, and Marc
  Pollefeys.
\newblock Real-time view correction for mobile devices.
\newblock {\em IEEE Transactions on Visualization and Computer Graphics}, 2017.

\bibitem{schops2017large}
Thomas Sch{\"o}ps, Torsten Sattler, Christian H{\"a}ne, and Marc Pollefeys.
\newblock Large-scale outdoor 3d reconstruction on a mobile device.
\newblock {\em Computer Vision and Image Understanding}, 2017.

\bibitem{Schoeps-et-al-TPAMI-2020}
T. {Schöps}, T. {Sattler}, and M. {Pollefeys}.
\newblock {SurfelMeshing}: Online surfel-based mesh reconstruction.
\newblock {\em IEEE Transactions on Pattern Analysis and Machine Intelligence},
  2019.

\bibitem{seitz1999voxelcoloring}
Steven~M Seitz and Charles~R Dyer.
\newblock Photorealistic scene reconstruction by voxel coloring.
\newblock {\em International Journal of Computer Vision}, 1999.

\bibitem{Sitzmann-et-al-CVPR-2019}
Vincent Sitzmann, Justus Thies, Felix Heide, Matthias Nie{\ss}ner, Gordon
  Wetzstein, and Michael Zollhofer.
\newblock {DeepVoxels}: Learning persistent 3d feature embeddings.
\newblock In {\em Proc.~International Conference on Computer Vision and Pattern
  Recognition (CVPR)}, 2019.

\bibitem{Sitzmann-et-al-NIPS-2019}
Vincent Sitzmann, Michael Zollh{\"o}fer, and Gordon Wetzstein.
\newblock Scene representation networks: Continuous 3d-structure-aware neural
  scene representations.
\newblock In {\em Proc.~Neural Information Processing Systems (NeurIPS)}, 2019.

\bibitem{srinivasan2019pushing}
Pratul~P Srinivasan, Richard Tucker, Jonathan~T Barron, Ravi Ramamoorthi, Ren
  Ng, and Noah Snavely.
\newblock Pushing the boundaries of view extrapolation with multiplane images.
\newblock In {\em Proc.~International Conference on Computer Vision and Pattern
  Recognition (CVPR)}, 2019.

\bibitem{steinbrucker2013large}
Frank Steinbrucker, Christian Kerl, and Daniel Cremers.
\newblock Large-scale multi-resolution surface reconstruction from rgb-d
  sequences.
\newblock In {\em Proc.~International Conference on Computer Vision (ICCV)},
  2013.

\bibitem{replica19arxiv}
Julian Straub, Thomas Whelan, Lingni Ma, Yufan Chen, Erik Wijmans, Simon Green,
  Jakob~J. Engel, Raul Mur-Artal, Carl Ren, Shobhit Verma, Anton Clarkson,
  Mingfei Yan, Brian Budge, Yajie Yan, Xiaqing Pan, June Yon, Yuyang Zou,
  Kimberly Leon, Nigel Carter, Jesus Briales, Tyler Gillingham, Elias Mueggler,
  Luis Pesqueira, Manolis Savva, Dhruv Batra, Hauke~M. Strasdat, Renzo~De
  Nardi, Michael Goesele, Steven Lovegrove, and Richard Newcombe.
\newblock The {R}eplica dataset: A digital replica of indoor spaces.
\newblock {\em arXiv preprint arXiv:1906.05797}, 2019.

\bibitem{stutz2018learning}
David Stutz and Andreas Geiger.
\newblock Learning 3d shape completion from laser scan data with weak
  supervision.
\newblock In {\em Proc.~International Conference on Computer Vision and Pattern
  Recognition (CVPR)}, 2018.

\bibitem{szeliski1998stereo}
Richard Szeliski and Polina Golland.
\newblock Stereo matching with transparency and matting.
\newblock In {\em Proc.~International Conference on Computer Vision (ICCV)},
  1998.

\bibitem{Takai-et-al-3DPVT-2010}
Takeshi Takai, Adrian Hilton, and Takashi Mastuyama.
\newblock Harmonised texture mapping.
\newblock In {\em Proc.~International Conference on 3D Vision (3DV)}, 2010.

\bibitem{Tarini-et-al-SIGGRAPH-Course-2017}
Marco Tarini, Cem Yuksel, and Silvain Lefebvre.
\newblock Rethinking texture mapping.
\newblock In {\em ACM SIGGRAPH 2017 Courses}, 2017.

\bibitem{tewari2020state}
A. Tewari, O. Fried, J. Thies, V. Sitzmann, S. Lombardi, K. Sunkavalli, R.
  Martin-Brualla, T. Simon, J. Saragih, M. Nießner, R. Pandey, S. Fanello, G.
  Wetzstein, J.-Y. Zhu, C. Theobalt, M. Agrawala, E. Shechtman, D.~B Goldman,
  and M. Zollhöfer.
\newblock State of the art on neural rendering.
\newblock {\em Computer Graphics Forum}, 2020.

\bibitem{Theobalt-et-al-TVCG-2007}
Christian Theobalt, Naveed Ahmed, Hendrik P.~A. Lensch, Marcus~A. Magnor, and
  Hans-Peter Seidel.
\newblock Seeing people in different light-joint shape, motion, and reflectance
  capture.
\newblock {\em IEEE Transactions on Visualization and Computer Graphics}, 2007.

\bibitem{thies2019neural}
Justus Thies, Michael Zollh{\"o}fer, and Matthias Nie{\ss}ner.
\newblock Deferred neural rendering: Image synthesis using neural textures.
\newblock {\em ACM Transactions on Graphics}, 2019.

\bibitem{thies2020imageguided}
Justus Thies, Michael Zollh{\"o}fer, Christian Theobalt, Marc Stamminger, and
  Matthias Nie{\ss}ner.
\newblock Image-guided neural object rendering.
\newblock In {\em Proc.~International Conference on Learning Representations
  (ICLR)}, 2020.

\bibitem{Tsiminaki-et-al-BMVC-2019}
Vagia Tsiminaki, Wei Dong, Martin~R. Oswald, and Marc Pollefeys.
\newblock Joint multi-view texture super-resolution and intrinsic
  decomposition.
\newblock In {\em Proc.~of the British Machine and Vision Conference (BMVC)},
  page~15. {BMVA} Press, 2019.

\bibitem{Tsiminaki-et-al-CVPR-2014}
Vagia Tsiminaki, Jean-S{\'e}bastien Franco, and Edmond Boyer.
\newblock High resolution 3d shape texture from multiple videos.
\newblock In {\em Proc.~International Conference on Computer Vision and Pattern
  Recognition (CVPR)}, 2014.

\bibitem{Waechter-et-al-ECCV-14}
Michael Waechter, Nils Moehrle, and Michael Goesele.
\newblock Let there be color! large-scale texturing of 3d reconstructions.
\newblock In {\em Proc.~European Conference on Computer Vision (ECCV)}, 2014.

\bibitem{wang2019real}
Kaixuan Wang, Fei Gao, and Shaojie Shen.
\newblock Real-time scalable dense surfel mapping.
\newblock In {\em IEEE International Conference on Robotics and Automation},
  2019.

\bibitem{wang2018pixel2mesh}
Nanyang Wang, Yinda Zhang, Zhuwen Li, Yanwei Fu, Wei Liu, and Yu-Gang Jiang.
\newblock Pixel2mesh: Generating 3d mesh models from single rgb images.
\newblock In {\em Proc.~European Conference on Computer Vision (ECCV)}, 2018.

\bibitem{wang2004image}
Zhou Wang, Alan~C Bovik, Hamid~R Sheikh, and Eero~P Simoncelli.
\newblock Image quality assessment: from error visibility to structural
  similarity.
\newblock {\em IEEE Transactions on Image Processing}, 2004.

\bibitem{weder2020routedfusion}
Silvan Weder, Johannes~L Sch{\"o}nberger, Marc Pollefeys, and Martin~R Oswald.
\newblock {RoutedFusion}: Learning real-time depth map fusion.
\newblock In {\em Proc.~International Conference on Computer Vision and Pattern
  Recognition (CVPR)}, 2020.

\bibitem{weder2020neuralfusion}
Silvan Weder, Johannes~L Sch{\"o}nberger, Marc Pollefeys, and Martin~R Oswald.
\newblock {NeuralFusion}: Online depth fusion in latent space.
\newblock In {\em Proc.~International Conference on Computer Vision and Pattern
  Recognition (CVPR)}, 2021.

\bibitem{whelan2015real}
Thomas Whelan, Michael Kaess, Hordur Johannsson, Maurice Fallon, John~J
  Leonard, and John McDonald.
\newblock Real-time large-scale dense rgb-d slam with volumetric fusion.
\newblock {\em International Journal of Robotics Research}, 2015.

\bibitem{whelan2015elasticfusion}
Thomas Whelan, Renato~F. Salas{-}Moreno, Ben Glocker, Andrew~J. Davison, and
  Stefan Leutenegger.
\newblock Elasticfusion: Real-time dense {SLAM} and light source estimation.
\newblock {\em International Journal of Robotics Research}, 2016.

\bibitem{wood2000surface}
Daniel~N Wood, Daniel~I Azuma, Ken Aldinger, Brian Curless, Tom Duchamp,
  David~H Salesin, and Werner Stuetzle.
\newblock Surface light fields for 3d photography.
\newblock In {\em ACM Transactions on Graphics (Proc.~SIGGRAPH)}, 2000.

\bibitem{Worrall-et-al-ICCV-2017}
Daniel~E Worrall, Stephan~J Garbin, Daniyar Turmukhambetov, and Gabriel~J
  Brostow.
\newblock Interpretable transformations with encoder-decoder networks.
\newblock In {\em Proc.~International Conference on Computer Vision (ICCV)},
  2017.

\bibitem{wu2016learning}
Jiajun Wu, Chengkai Zhang, Tianfan Xue, Bill Freeman, and Josh Tenenbaum.
\newblock Learning a probabilistic latent space of object shapes via 3d
  generative-adversarial modeling.
\newblock In {\em Proc.~Neural Information Processing Systems (NeurIPS)}, 2016.

\bibitem{Xu-et-al-NIPS-2019}
Qiangeng Xu, Weiyue Wang, Duygu Ceylan, Radom{\'{\i}}r Mech, and Ulrich
  Neumann.
\newblock {DISN:} deep implicit surface network for high-quality single-view 3d
  reconstruction.
\newblock In {\em Proc.~Neural Information Processing Systems (NeurIPS)}, 2019.

\bibitem{Yuksel-et-al-TOG-2010}
Cem Yuksel, John Keyser, and Donald~H. House.
\newblock Mesh colors.
\newblock {\em ACM Transactions on Graphics}, 2010.

\bibitem{yuksel2019rethinking}
Cem Yuksel, Sylvain Lefebvre, and Marco Tarini.
\newblock Rethinking texture mapping.
\newblock In {\em Computer Graphics Forum}, 2019.

\bibitem{zeng2012memory}
Ming Zeng, Fukai Zhao, Jiaxiang Zheng, and Xinguo Liu.
\newblock A memory-efficient kinectfusion using octree.
\newblock In {\em Proc.~International Conference on Computational Visual
  Media}, 2012.

\bibitem{zeng2013octree}
Ming Zeng, Fukai Zhao, Jiaxiang Zheng, and Xinguo Liu.
\newblock Octree-based fusion for realtime 3d reconstruction.
\newblock {\em Graphical Models}, 2013.

\bibitem{zhang2020neural}
Xiuming Zhang, Sean Fanello, Yun-Ta Tsai, Tiancheng Sun, Tianfan Xue, Rohit
  Pandey, Sergio Orts-Escolano, Philip Davidson, Christoph Rhemann, Paul
  Debevec, et~al.
\newblock Neural light transport for relighting and view synthesis.
\newblock {\em ACM Transactions on Graphics}, 2021.

\bibitem{Zollhofer-et-al-TOG-2015}
Michael Zollh{\"o}fer, Angela Dai, Matthias Innmann, Chenglei Wu, Marc
  Stamminger, Christian Theobalt, and Matthias Nie{\ss}ner.
\newblock {Shading-based Refinement on Volumetric Signed Distance Functions}.
\newblock {\em ACM Transactions on Graphics}, 2015.

\end{thebibliography}
